\documentclass[runningheads]{llncs}

\usepackage{eccv}

\usepackage{eccvabbrv}

\usepackage{graphicx}
\usepackage{booktabs}
\usepackage{multirow}
\usepackage{amsmath}
\usepackage{xcolor}
\definecolor{brightgray}{RGB}{220,220,220}
\usepackage{colortbl}
\usepackage{comment}

\newcommand{\Post}{{\textbf{\text{SaLS}}}}
\newcommand{\Norm}{{\textbf{\text{ZS-Norm}}}}
\newcommand{\Pen}{{\textbf{\text{Penalty}}}}

\usepackage[accsupp]{axessibility}  

\def\vl{{\bm{l}}}

\def\vt{{\bm{t}}}
\def\vw{{\bm{w}}}
\def\vx{{\bm{x}}}
\def\vy{{\bm{y}}}
\def\vz{{\bm{z}}}

\def\VP{{\bm{P}}}
\def\VY{{\bm{Y}}}

\def\real{{\mathbb{R}}}
\newcommand{\appen}{\textcolor{red}{Appendix}\xspace}

\usepackage{hyperref}

\usepackage{orcidlink}
\usepackage{bm,upgreek}

\begin{document}

\title{Robust Calibration of \\ Large Vision-Language Adapters} 

\titlerunning{Robust Calibration of Large Vision-Language Adapters}

\author{Balamurali Murugesan \orcidlink{0000-0002-3002-5845} \and
Julio Silva-Rodríguez \orcidlink{0000-0002-9726-9393} \and \\
Ismail Ben Ayed \orcidlink{0000−0002−9668−8027} \and Jose Dolz \orcidlink{0000-0002-2436-7750}}

\authorrunning{B.~Murugesan et al.}

\institute{ETS Montreal, Canada \\ \email{balamurali.murugesan.1@ens.etsmtl.ca}}

\maketitle

\begin{abstract}
  This paper addresses the critical issue of miscalibration in CLIP-based model adaptation, particularly in the challenging scenario of out-of-distribution (OOD) samples, which has been overlooked in the existing literature on CLIP adaptation. We empirically demonstrate that popular CLIP adaptation approaches, such as Adapters, Prompt Learning, and Test-Time Adaptation, substantially degrade the calibration capabilities of the zero-shot baseline in the presence of distributional drift. We identify the increase in logit ranges as the underlying cause of miscalibration of CLIP adaptation methods, contrasting with previous work on calibrating fully-supervised models. Motivated by these observations, we present a simple and model-agnostic solution to mitigate miscalibration, by scaling the logit range of each sample to its zero-shot prediction logits. We explore three different alternatives to achieve this, which can be either integrated during adaptation or directly used at inference time. Comprehensive experiments on popular OOD classification benchmarks demonstrate the effectiveness of the proposed approaches in mitigating miscalibration while maintaining discriminative performance, whose improvements are consistent across the three families of these increasingly popular approaches. The code is publicly available at: \url{https://github.com/Bala93/CLIPCalib} \ .
  
\end{abstract}

\begin{figure*}[h!]
     \begin{subfigure}[b]{0.24\linewidth}
         \centering
         \includegraphics[width=\linewidth]{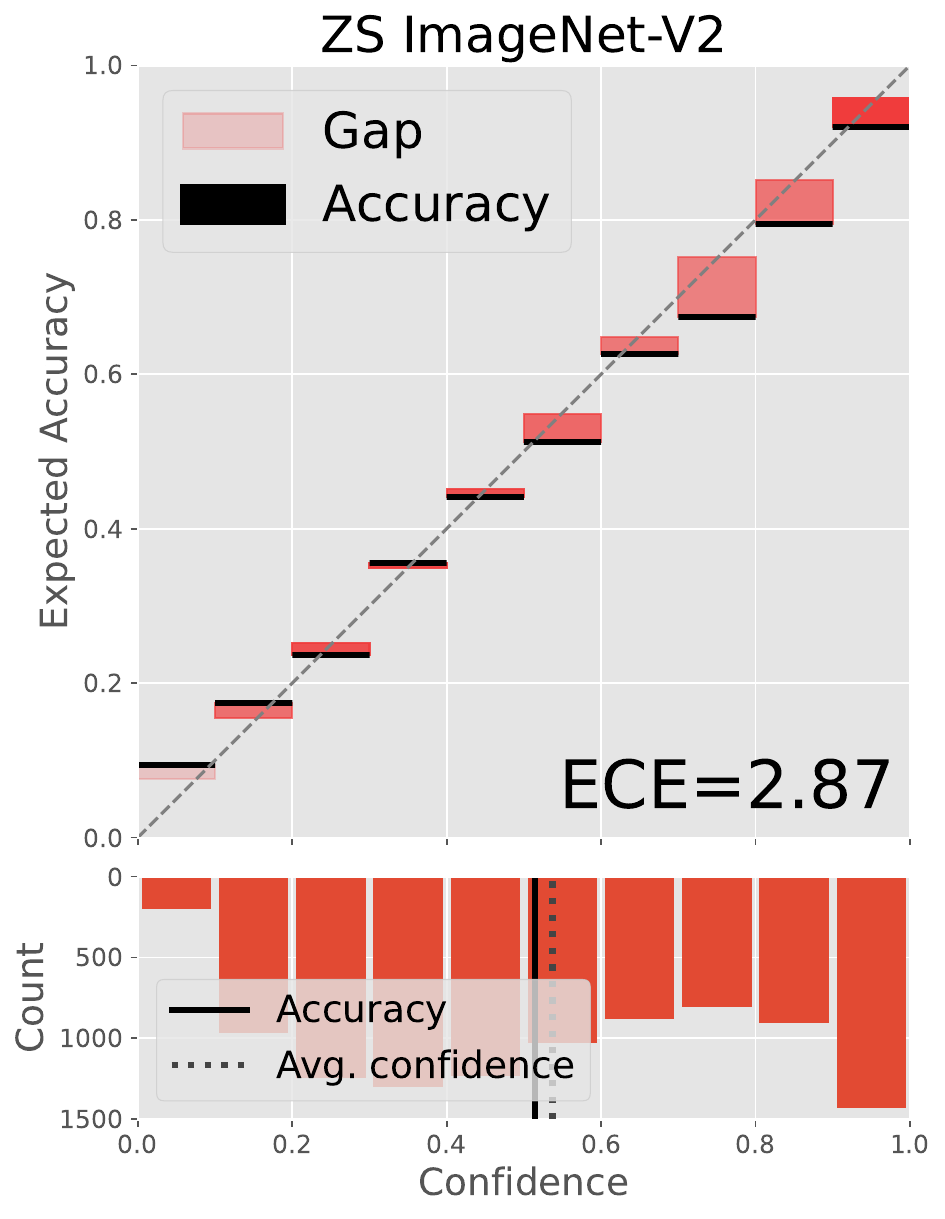}
     \end{subfigure}
     \begin{subfigure}[b]{0.24\linewidth}
         \centering
         \includegraphics[width=\linewidth]{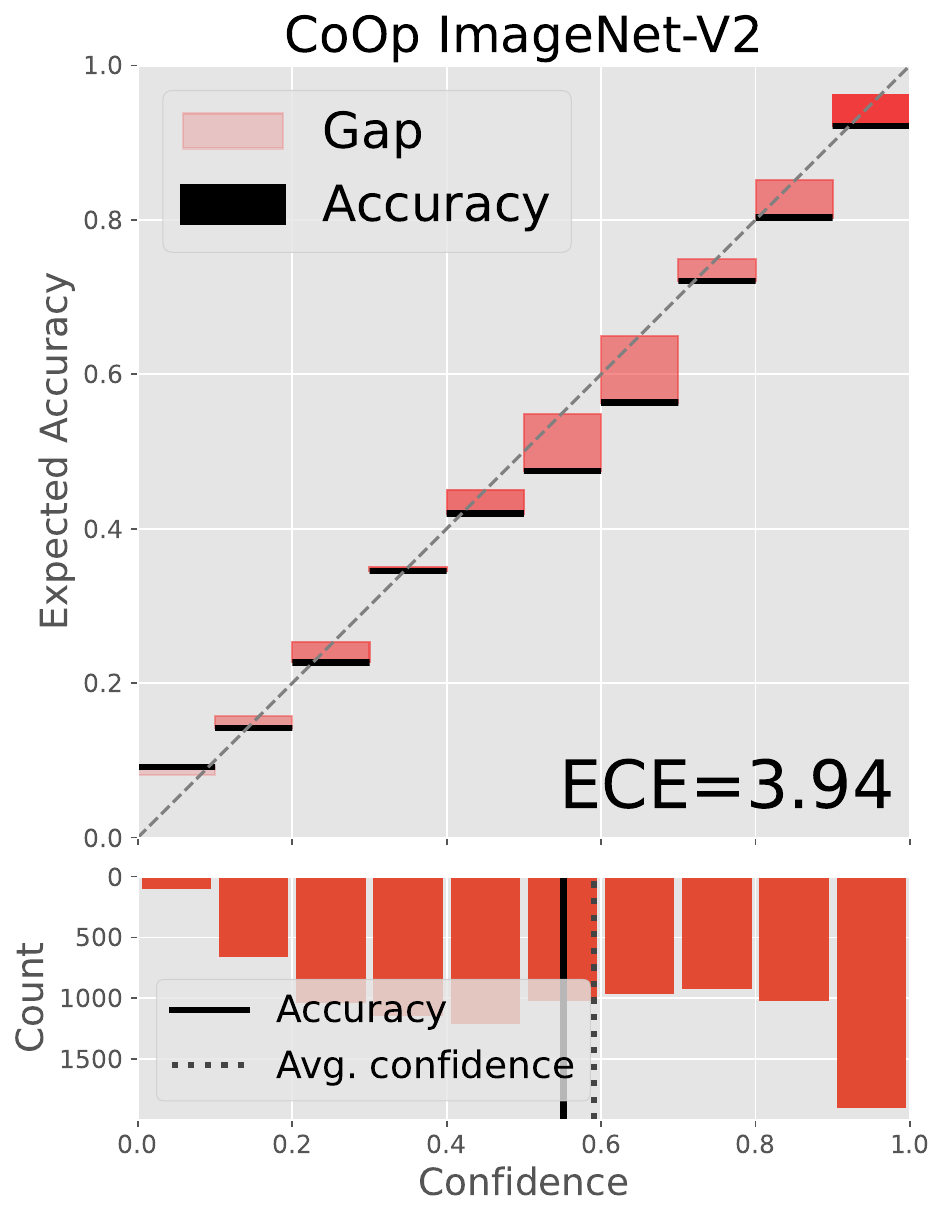}
     \end{subfigure}
     \begin{subfigure}[b]{0.24\linewidth}
         \centering
         \includegraphics[width=\linewidth]{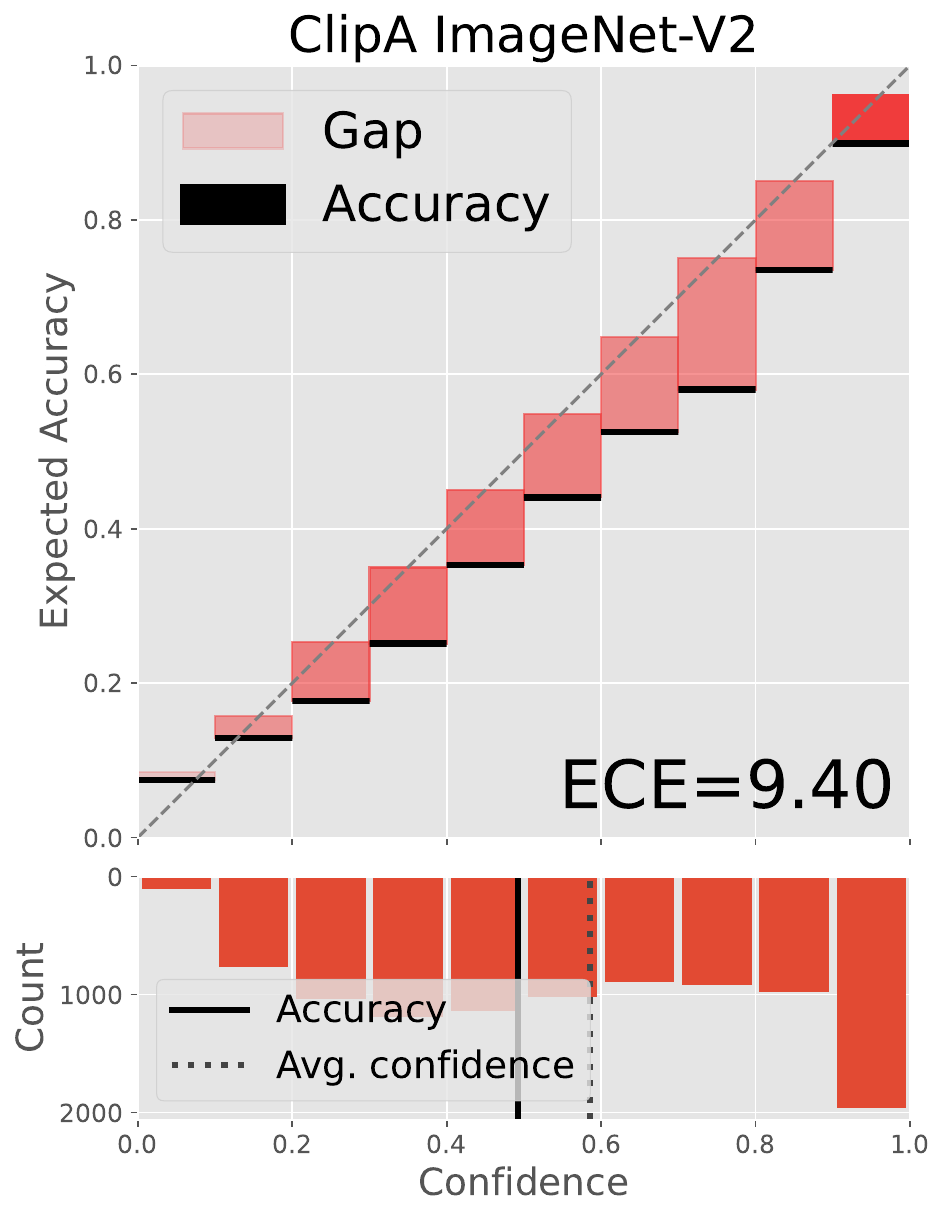}
     \end{subfigure}
     \begin{subfigure}[b]{0.24\linewidth}
         \centering
         \includegraphics[width=\linewidth]{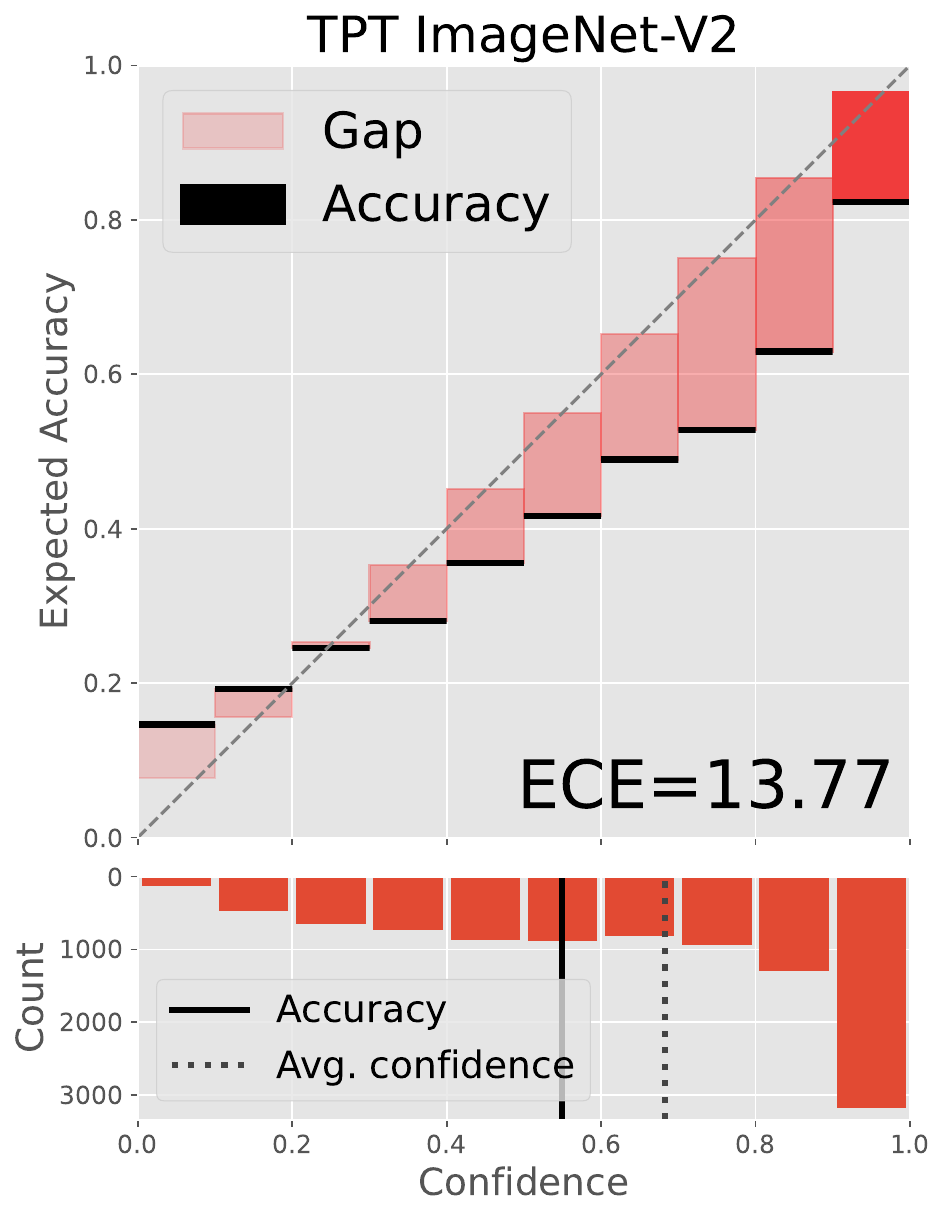}
     \end{subfigure}
     \begin{subfigure}[b]{0.24\linewidth}
         \centering
         \includegraphics[width=\linewidth]{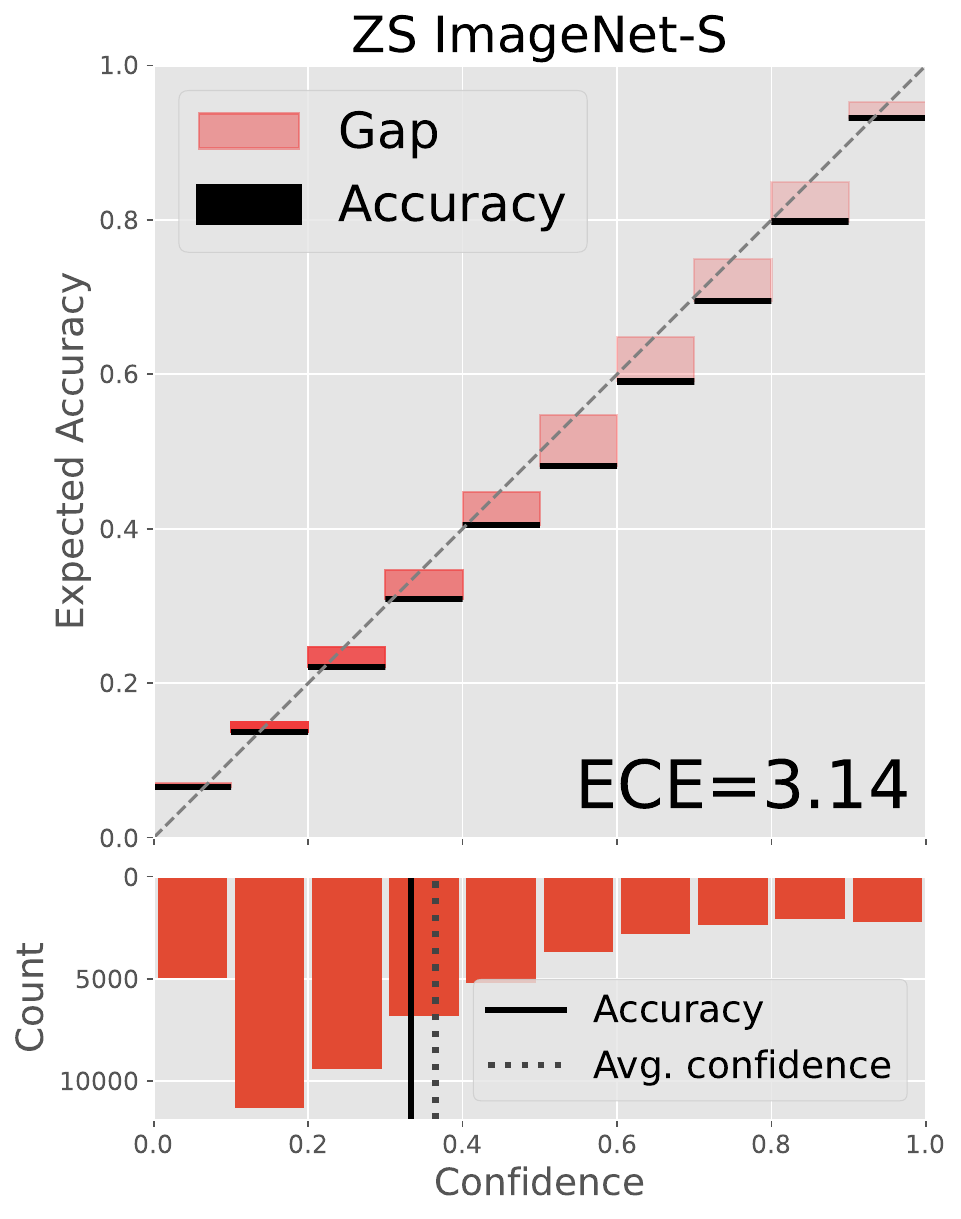}
     \end{subfigure}
     \begin{subfigure}[b]{0.24\linewidth}
         \centering
         \includegraphics[width=\linewidth]{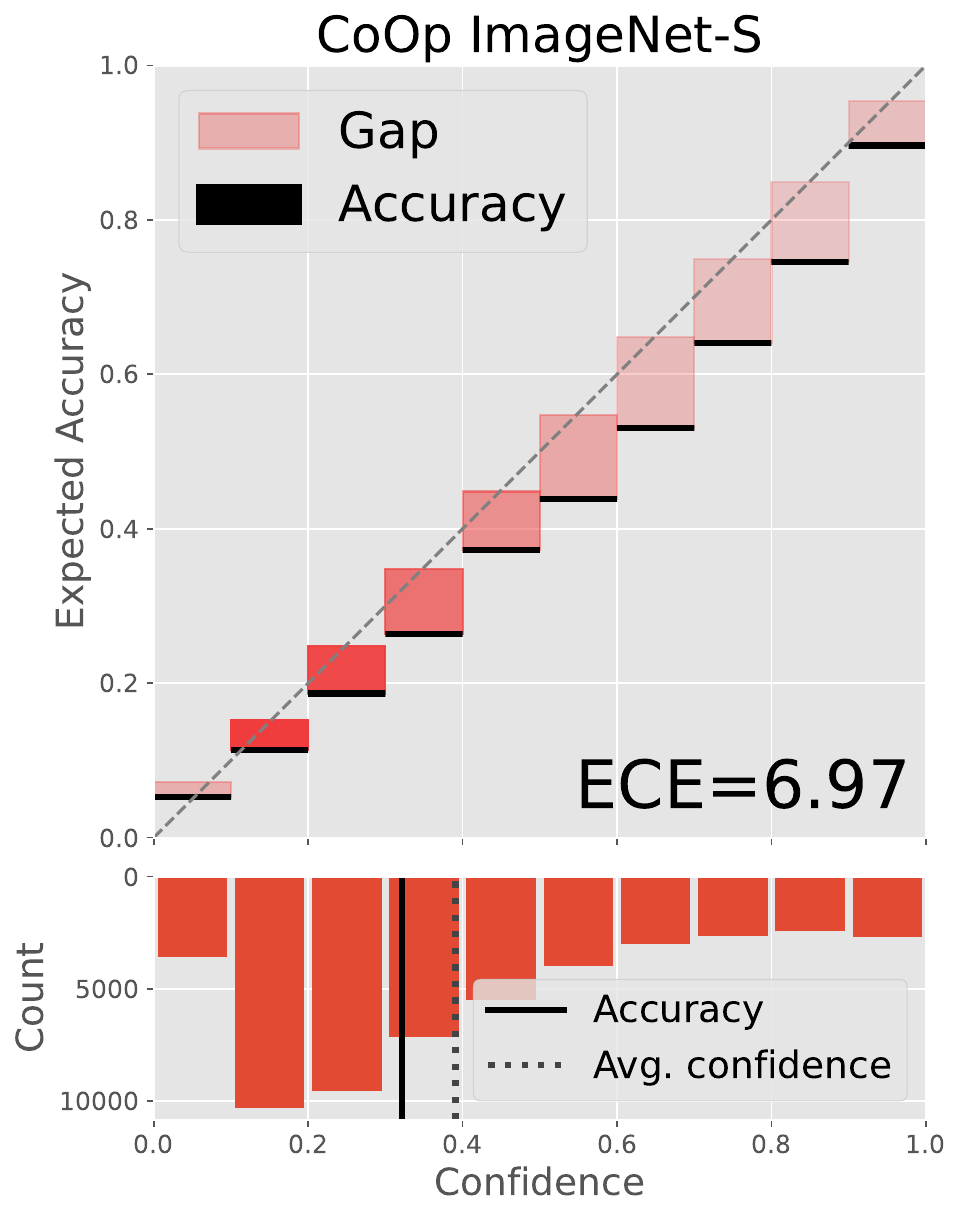}
     \end{subfigure}
     \begin{subfigure}[b]{0.24\linewidth}
         \centering
         \includegraphics[width=\linewidth]{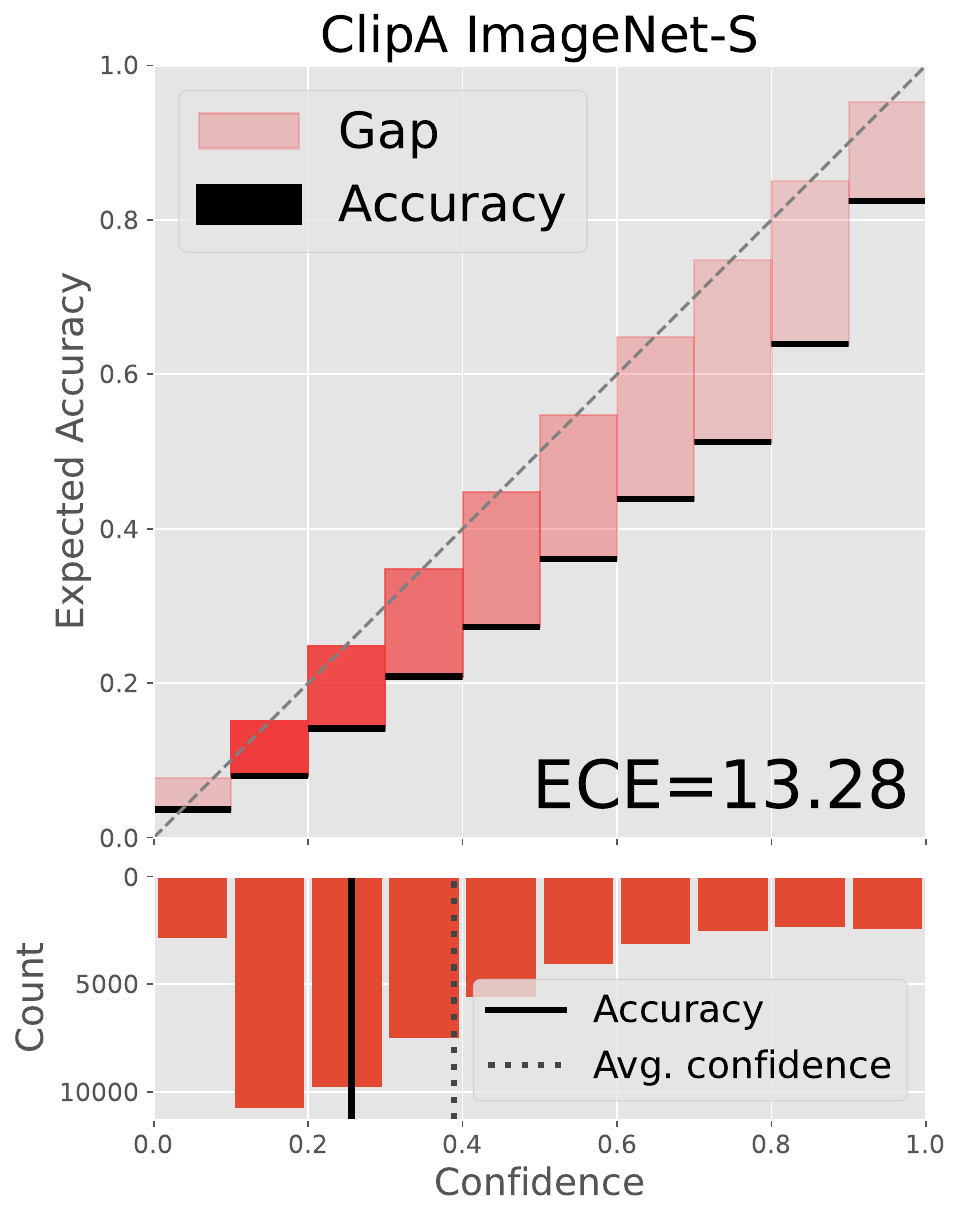}
     \end{subfigure}
     \begin{subfigure}[b]{0.24\linewidth}
         \centering
         \includegraphics[width=\linewidth]{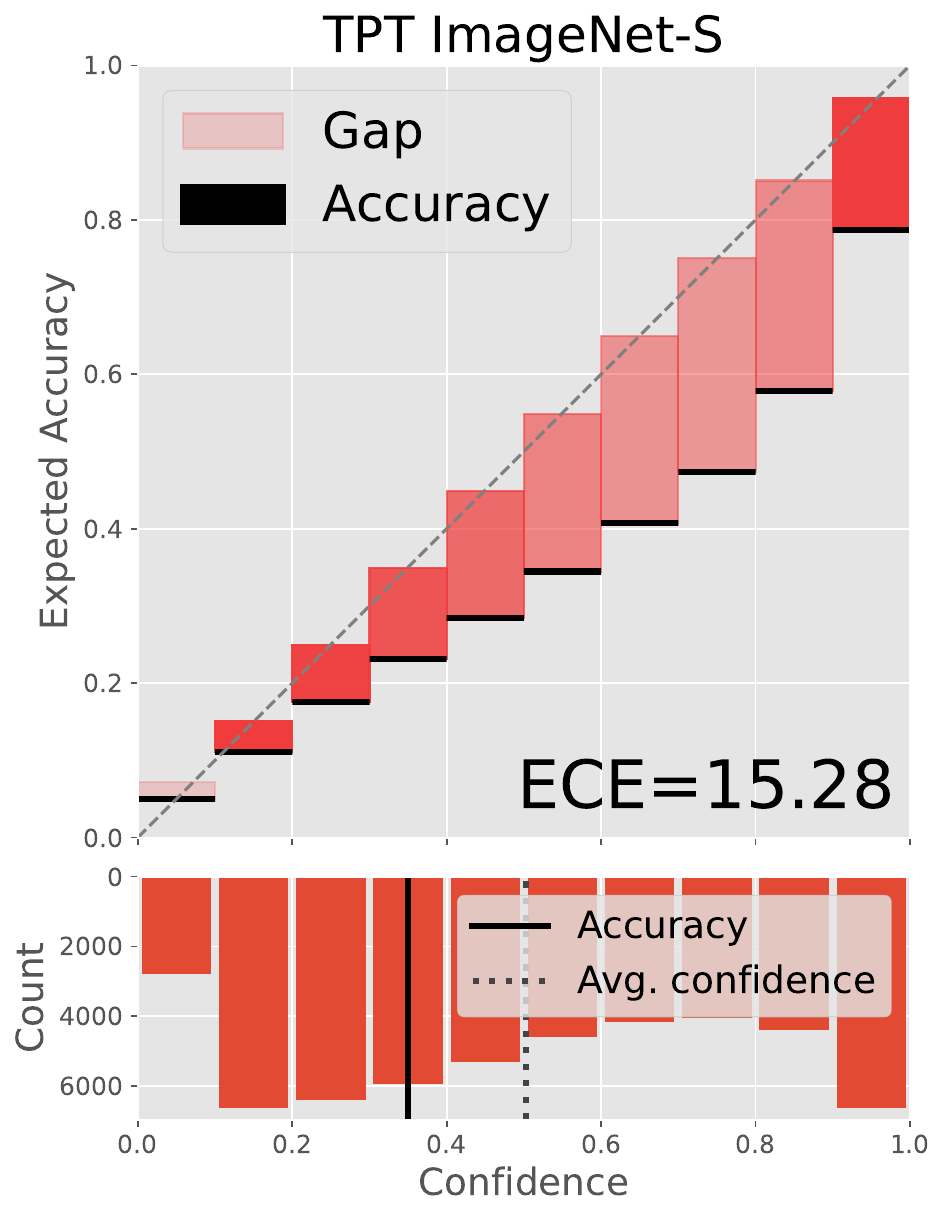}
     \end{subfigure}
    \caption{\textbf{CLIP-based adaptation methods are severely miscalibrated on Out-of-distribution (OOD) samples.} Three families of popular approaches to adapt CLIP under different scenarios, i.e., Prompt Learning (CoOp \cite{zhou2022coop}), Adapters (Clip-Ad \cite{gao2021clip}) and Test-time prompt tuning  (TPT \cite{tpt}), significantly degrade the miscalibration of the zero-shot baseline, despite improving its discriminative performance.}
    \label{fig:miscalibration}
\end{figure*}

\section{Introduction}
\label{sec:intro}

Deep learning is undergoing a paradigm shift with the emergence of pre-training large-scale language-vision models, such as CLIP \cite{radford2021learning}. These models, and more particularly the variants integrating vision transformers, have demonstrated impressive generalization capabilities in visual recognition tasks, yielding exceptional zero-shot and few-shot performance. 
Nevertheless, in a dynamic and evolving open world, machine learning applications inevitably encounter the challenge of out-of-distribution (OOD) data, which typically hinders the scalability of these models to new domains. Existing literature based on CLIP faces this scenario with different solutions to exacerbate robustness. In particular, freezing the entire vision backbone to re-use these generalizable features has been a popular choice, especially in the low-data regime \cite{gao2021clip,zhou2022coop}. Thus, CLIP adaptation during training typically resorts to Adapters \cite{gao2021clip,zhang2021tip} or Prompt Learning \cite{zhou2022coop,zhou2022cocoop} strategies, which leverage a few labeled samples to adapt the model with the hope that it will generalize properly to unseen related-domains. Furthermore, a more challenging scenario consists of adapting the model during inference without any access to labeled data, where a prevalent method is Test-Time Prompt Tuning (TPT) \cite{tpt}. 

While these strategies have further improved the discriminative performance of a zero-shot baseline, we have observed that the accuracy of the uncertainty estimates of the predictions, i.e., calibration, is significantly degraded (see \cref{fig:miscalibration}), regardless of the family of adaptation models or setting. Thus, after adaptation, model predictions are often over-confident, even if they are wrong. This represents a major concern, as inaccurate uncertainty estimates can carry serious implications in safety-critical applications, such as healthcare, where CLIP is emerging as a popular strategy \cite{liu2023clip,liang2022effective}. Nevertheless, despite its importance, the miscalibration issue has been overlooked in the CLIP adaptation literature. 

Motivated by these observations, this paper addresses this critical issue, which has been disregarded in current literature. Indeed, few-shot adaptation strategies, notably Prompt Learning and Adapters, are attracting wide attention recently, with an unprecedented surge in the number of methods proposed \cite{liu2023clip,zhang2021tip,yu2023task,silva2023closer,hu2021lora,zhou2022coop,zhou2022cocoop,kgcoop23,khattak2023maple}, albeit being a relatively recent research topic. Nevertheless, the main focus of this growing literature has been on improving the discriminative power of adapted models. Thus, given their increasing popularity, and quick adoption in real-world safety-critical problems, we believe that assessing the calibration performance of CLIP adaptation strategies in OOD scenarios is of paramount importance to deploy not only high-performing but also reliable models. 
We can summarize our contributions as follows:
 
\begin{enumerate}
    \item We empirically demonstrate that popular CLIP adaptation strategies, such as Adapters, Prompt Learning, and Test-Time Prompt Tuning, substantially degrade the calibration capabilities of the zero-shot baseline in the presence of distributional drift.
    \item For these adaptation strategies, we expose that the underlying cause of miscalibration is, in fact, the increase of the logit ranges. This contrasts with recent work in calibrating fully-supervised models \cite{wei2022mitigating}, which suggests that the inherent cause of miscalibration is the increase of its norm instead, due to the standard cross-entropy loss used for training. 
    \item Based on these observations, we present a simple, and model-agnostic solution, which consists in scaling the logit range of each sample based on the zero-shot logits. We further present several alternatives to accommodate our solution, which can be implemented either at training or inference time. 
    \item Comprehensive experiments on popular OOD classification benchmarks empirically demonstrate the effectiveness of our approaches to reduce the miscalibration error, while keeping the discriminative performance. 
\end{enumerate}

\section{Related Work}
\label{sec:rw}

\subsection{Vision language models}

Text-driven pre-training of image representation, so-called vision-language models (VLMs) is revolutionizing the paradigm of transfer learning. These models can integrate massive web-scrabbled data sources thus learning robust feature representations. In particular, models such as CLIP \cite{radford2021learning} or ALIGN \cite{jia2021scaling} train joint multi-modal embedding spaces via contrastive learning of paired images and text, using dual encoder architectures. Such strong vision-language alignment has demonstrated robust open-vocabulary zero-shot generalization capabilities \cite{radford2021learning,zhai2022lit}. Given such potential, transferring pre-trained VLMs to a wide variety of tasks is gaining increasing popularity. Nevertheless, this process faces particular challenges. First, large-scale pre-training usually involves also scaling network sizes, which is a computational bottleneck for low-resource adaptation scenarios. Second, recent attempts to fine-tune VLMs have demonstrated a deterioration of their robustness against domain drifts \cite{LPFT,WiSE}, especially when available data is limited. Thus, an emerging core of recent literature is focusing on novel alternatives to overcome these limitations. More concretely, freezing the pre-trained backbone, and reusing such features by training a small set of parameters, via Prompt Learning \cite{zhou2022coop,zhou2022cocoop,kgcoop23,zhu2023prompt,khattak2023maple}, or black-box Adapters \cite{gao2021clip,zhang2021tip,yu2023task,silva2023closer,ouali2023black,li2024graphadapter,zhang2023prompt}, is getting increasing attention. 

\subsection{Prompt based learning}

CLIP models have shown encouraging results by hand-crafting personalized text descriptions of the target visual representation \cite{Menon2023}. The automatizing of this cumbersome process raises the concept of Prompt Learning (PL) \cite{zhou2022coop}, a family of methods to adapt CLIP that inserts a set of continuous learnable tokens in the original text prompt at the input of the VLM language encoder. While the CLIP model remains frozen, PL optimizes the most discriminative text input, given a few-shot support set \cite{zhou2022coop,zhou2022cocoop,khattak2023maple,zhu2023prompt}. CoOP \cite{zhou2022coop} represents one of the initial attempts to study the effect of prompt tuning on different tasks, and proposed to learn the prompt's context words. CoCoOP \cite{zhou2022cocoop}, on the other hand, designed a simple network to predict the input text prompt through image features, as CoOP failed to match the zero-shot performance on generic tasks. TPT \cite{tpt} extends PL to address time-test adaptation scenarios by updating the prompt for a batch with original and augmented samples through entropy minimization.

\subsection{Black-box Adapters}
Prompt Learning involves using the CLIP's encoder throughout the entire training process as the backpropagation of the gradient has to pass through it to update the prompts, which results in large computational constraints \cite{gao2021clip}. Adapter-based techniques provide an alternative to Prompt Learning for aligning to downstream tasks, leveraging uniquely pre-computed features with minimal additional parameters. A base version of such methods involves training a linear classifier via logistic regression, typically referred to as Linear Probing \cite{radford2021learning}. Nevertheless, leveraging only the vision features does not fully exploit the potential of VLMs. To this end, several methods have proposed enhanced Adapters, which further rely on zero-shot text-driven class-wise prototypes. In particular, Clip-Adapter \cite{gao2021clip} introduced additional fully connected layers and operated on the vision or language branch through residual style feature combination. Training-free methods such as Tip-Adapter \cite{zhang2021tip} resorted to a key-value cache model based on the available few-shot supports. Likewise, TaskRes \cite{yu2023task} introduced additional learning parameters and applied a residual modification of the text representation, which led to a better initialization when learning from few-shot supervision. More recently, \cite{silva2023closer} provided a wider look at the coupling of vision and text features in such Adapters, by pointing out that these methods largely build up their improved performance on initializing the logistic classifier weights with the zero-shot prototypes, proposing a simple solution, coined CLAP, for a better distillation of such prototypes.

\subsection{Model calibration}
Calibrating the confidence of deep learning models is paramount in developing reliable solutions, as the confidence is expected to correlate with correctness. Given the importance and the potential impact of miscalibration, a growing literature to address this issue has emerged in the last years. Post-processing techniques have been widely used to achieve calibration, wherein a linear transformation \cite{guo2017calibration, tomani2022parameterized, joy2023sample} is applied to the predicted logits before converting it to softmax. Nevertheless, an important limitation is that these transformations are obtained using held-out validation data, which is assumed to follow the same distribution as the test data, limiting their applicability in the presence of domain drifts \cite{ovadia2019can}. A popular alternative consists in calibrating the networks at training time. This can be achieved by incorporating explicit penalties that either penalize overconfident softmax predictions \cite{pereyra2017regularizing,larrazabal2021maximum,cheng2022calibrating,park2023acls} or encourage small logit differences \cite{murugesan2023trust,liu2022devil,liu2023class}. Furthermore, \cite{muller2019does,mukhoti2020calibrating} demonstrated that popular classification losses, such as Focal Loss \cite{lin2017focal} or Label Smoothing \cite{szegedy2016rethinking}, integrate an implicit term that maximizes the entropy of the network predictions, thus favoring low-confidence models. Other works to improve the accuracy of the uncertainty estimates during training include the use of MixUp \cite{thulasidasan2019mixup,zhang2022and}, or enforcing a constant vector norm on the logits \cite{wei2022mitigating}, among others. Nevertheless, all these methods have been proposed in the context of fully-supervised models, and the calibration of Prompt Learning and Adapter-based methods for CLIP remains unexplored in the literature. 

\section{Background}
\label{preliminaries}

\subsection{CLIP Zero-Shot Classification}

CLIP \cite{radford2021learning} is a large vision-language model, trained via contrastive learning to produce visual representations from images $\vx$ paired with their associated text descriptions $T$. To do so, CLIP consists of an image encoder $\boldsymbol{\theta}$ and a text encoder $\boldsymbol{\phi}$. This generates the corresponding vision $\vz \in \real^d$ and class text $\vw_k \in \real^d$ embeddings, which are typically projected into an $\ell_{2}$-normalized shared embedding space. Given a new task consisting in visually discriminating between $K$ categories, the set containing all the text embeddings for all the $K$ classes can be denoted as $\mathcal{W}=\{\vw_k\}_k^K$, with $\vw_k=\boldsymbol{\phi}(\textsc{``A photo of a [class$_k$]''})$. At inference, this learning paradigm enables zero-shot prediction. More concretely, for a given set of $K$ classes, and an ensemble of $N$ different prompts per category, we can generate the set of available prompts as $\mathcal{T}=\{\{T_{n,k}\}_{n=1}^N\}_{k=1}^K\}$. Then, a popular strategy \cite{radford2021learning,gao2021clip,WiSE} consists in obtaining a class zero-shot prototype, which is computed as $\vt_k=\frac{1}{N}\sum^N_{n=1}\boldsymbol{\phi}(T_{n,k})$. Then, for a given test image, the zero-shot prediction, $\bm{p}=(p_k)_{1 \leq k \leq K}$,  can be obtained as:
\begin{equation}
\label{eq:clip}
p_k = \frac{\exp{(\vz^\top \vt_k/\tau)}}{\sum_{j=1}^K\exp{(\vz^\top \vt_j/\tau)}}
\end{equation}

\noindent where $\tau$ is a temperature hyperparameter, whose value is learned during training, and $\vz^\top \vt$ denotes the dot product operator\footnote{As vectors are $\ell_{2}$-normalized, the dot product between these two vectors is equivalent to their cosine similarity.}.

\subsection{Adaptation to novel tasks}

Let us now consider a support set that contains a few labeled samples $\mathcal{S}=\{(\vx_i,\vy_i)\}_{i=1}^S$, with $\vy \in \{0,1\}^K$ the ground truth vector associated with $\vx$. The vector of predicted logits of a given image $i$ is defined as 
$\vl_i=(l_{ik})_{1 \leq k \leq K}$. In Prompt Learning methods, such as CoOp \cite{zhou2022coop} or KgCoOp \cite{zhou2022cocoop}, the adaptation is done by modeling the input text $T_k$ of a given class $k$ as learnable continuous vectors. Thus, in contrast with zero-shot inference, where the resulting text embeddings are obtained as the mean over the different pre-defined prompts, in Prompt Learning these are optimized. To generate the logits, the learnable prompts are combined with the fixed visual embedding from the test image $i$, such that $l_{ik}=\vz_i^\top\vt_k / \tau$, which can then be integrated into \cref{eq:clip} to minimize the cross entropy loss over the few labeled shots. The family of methods commonly referred to as Adapters \cite{gao2021clip,zhang2021tip,yu2023task,silva2023closer} proceeds differently, and learns transformations over the visual and text embeddings, yielding the following logits $l_{ik}=\boldsymbol{\theta}_a(\vz_i,\vt_k, \tau)$, where $\boldsymbol{\theta}_a$ is the set of learnable parameters of the Adapter. A more challenging scenario consists in adapting the text prompts at inference, which is commonly referred to as test-time prompt tuning \cite{tpt}. As this setting does not include few-shot supports to adapt the prompts, the supervised cross-entropy objective is replaced by an unsupervised minimization of the Shannon entropy. Thus, regardless of the method selected, the objective is to optimize either $\vt_k$ (Prompt Learning and test-time prompt tuning) or $\boldsymbol{\theta}_a$ (Adapters) to minimize either the CE over the softmax predictions obtained from the few-shots, or the Shannon entropy on the test samples predictions at inference. 

\section{Constraining logits during adaptation}
\subsection{Impact of adaptation in logits}

To understand the impact on calibration of using the cross-entropy (CE) loss to adapt CLIP, let us decompose the logit vector $\vl$ into its Euclidean norm $\|\vl\|=\sqrt{l_1^2+...+l_K^2}$, (\textit{magnitude}) and its unit vector $\hat{\vl}$ (\textit{direction}), such that $\vl = \|\vl\| \hat{\vl}$. Considering now the \textit{magnitude} and \textit{direction} of the logit vector, the general form of the cross-entropy loss over a given support sample, using the softmax probabilities in \cref{eq:clip}, can be formulated as:

\begin{equation}
\label{eq:ce}
-\log \frac{\exp{(\|\vl_i\| \hat{l}_{ik})}}{\sum_{j=1}^{K}\exp{(\|\vl_i\| \hat{l}_{ij})}}
\end{equation}

\vspace{+2em}

This view of the cross-entropy implies 
that the direction of the logit vector $\hat{\vl}_i$ determines the predicted class of the image $i$. Thus, if the predicted category is incorrect, $\hat{\vl}_i$ will change to match the target class provided in the one-hot encoded label. Once the network prediction is correct, i.e., $y_i=\arg \max_j(l_{ij})$, the direction of the vector will remain unchanged. Nevertheless, the nature of the cross-entropy loss will favor higher softmax probabilities 
for the predicted class. Recent literature \cite{wei2022mitigating} suggests that this is achieved by increasing $\|\vl_i\|$, indicating that the miscalibration issue originates from the augmentation of the logit norm. 
Nevertheless, in what follows we refute this argument and advocate for \textbf{the increase of the logits range as the potential cause of miscalibration.}

\noindent \textbf{Proposition 1.} \textit{Let us consider the softmax cross entropy loss, where $\sigma(\cdot)$ denotes the softmax function. Assume that $\vl \geq \mathbf{0}$.
Then, for any scalar $a>0$, $\sigma_k(\vl)=\sigma_k(\vl+a) \, \forall k$, and $\|\vl+a \mathbf{1}\|>\|\vl\|$, where $\mathbf{1}$ denotes the vector of ones.}

Prop. 1 demonstrates that adding a strictly positive constant value $a \in \real_{++}$ to all the logits increases the norm of the vector $\vl$, but this does not necessarily lead to more confident predictions, whose probability scores remain unchanged.

\noindent \textbf{Proposition 2.} \textit{Let $R(\vl) = \max(\vl) - \min(\vl)$ denotes the range of logit vector $\vl$, where $\max(\vl)$ (respectively $\min(\vl)$) denotes the largest (respectively smallest) value among the elements of vector $\vl$. Then, for any given scalar $a>1$, and for $k=\arg \max_j(l_j)$, we have $\sigma_k(a \vl)>\sigma_k(\vl)$ and $R(a \vl)>R(\vl)$.}

\begin{figure}[h!]
    \centering
    \includegraphics[width=0.95\linewidth]{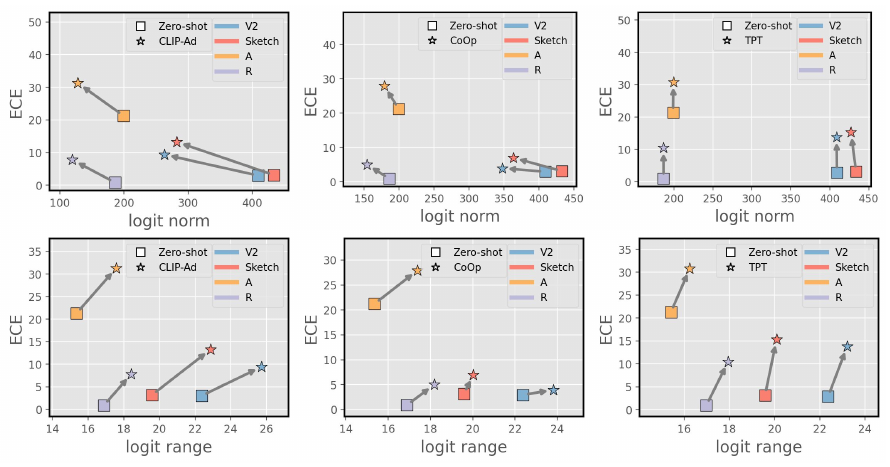}
    \caption{\textbf{Logit norm or logit range as the source of miscalibration?} These figures clearly show that when the calibration of the zero-shot (ZS) model is degraded, the logit norm of its predictions is reduced (\textit{top}), which discards an increase of the logit norm as the main cause for miscalibration. In contrast, there exists a correlation between the increase of the logit ranges and miscalibration (\textit{bottom}).}
    \label{fig:logit-ranges}
\end{figure}

The proofs of Propositions 1 and 2 are deferred to the \appen. From the above proposition we find that increasing the range of a given logit vector results in higher softmax probability values. Thus, contrary to the widely spread belief that increasing the logit norm hinders 
model calibration, we argue that this effect of \textit{logit distance magnification}, which yields higher softmax distributions, is a potential source of miscalibration\footnote{Note that the same reasoning applies to TPT \cite{tpt}, whose learning objective to adapt 
the CLIP baseline is to minimize the Shannon entropy of the softmax 
distribution.}. This explains why, even though adaptation of CLIP yields performance gains in terms of accuracy, adapted models are worse calibrated than a zero-shot baseline. Furthermore, this analysis is supported empirically by the observations depicted in \cref{fig:logit-ranges}, where can observe that, while calibration has been degraded in the adapted models, the logit norm of their predictions has substantially decreased.

\subsection{Our solution}
\label{methods}

From our previous analysis and empirical observations, we can derive that: \textit{i)} despite improving their classification performance, state-of-the-art strategies to adapt CLIP suffer from miscalibration, particularly compared to the original zero-shot predictions, and \textit{ii)} one of the main causes arises from the logit magnification issue introduced by the cross-entropy loss used during adaptation. 

In light of these findings, we propose a simple but effective solution that can alleviate the miscalibration issue in CLIP adaptation. More concretely, we propose to constraint the range of the logits during the training of a main objective $\mathcal{H}$, which results in the following constrained problem:

\begin{align}
\label{eq:ce-constrained}
&\text{minimize} \qquad \mathcal{H}(\VY,\VP) \nonumber \\
&\text{subject to} \qquad l^{\text{ZS-min}}_{i} \mathbf{1} \leq \vl_i \leq l^{\text{ZS-max}}_{i} \mathbf{1} \qquad \forall i \in \mathcal{D},
 \end{align}
where $\VY$ and $\VP$ are matrices containing the sample-wise ground-truth and softmax-prediction vectors for all the samples involved in the training, $l^{\text{ZS-min}}_{i}$ and $l^{\text{ZS-max}}_{i}$ are the min and max logit magnitudes of the zero-shot prediction for sample $\vx_i$. $\mathcal{D}$ denotes a given set of available samples. Furthermore, in the test-time prompt tuning setting, we simply need to replace $\VY$ by $\VP$ in \cref{eq:ce-constrained}. Directly solving the constrained problem in \cref{eq:ce-constrained} in the context of deep models is not trivial \cite{marquez2017imposing}, and Lagrangian-dual optimization has been typically avoided in modern deep networks involving millions of parameters. To address this issue, we propose several alternatives to approximate the constrained problem presented in \cref{eq:ce-constrained}, which are detailed below.

\subsection{Zero-shot logit normalization during training ($\Norm$)}
The constraint in the presented problem, i.e., $l^{\text{ZS-min}}_{i} \mathbf{1} \leq \vl_i \leq l^{\text{ZS-max}}_{i} \mathbf{1}, \forall i \in \mathcal{D}$, can be integrated into the main objective by transforming the logits before computing the CE loss over the support set samples (here $\mathcal{D}=\mathcal{S}$). More concretely, the modified learning objective can be defined as:

\begin{equation}
\label{eq:ce-logitnorm}
\mathcal{H}(\VY,\VP)=-\sum_{i\in \mathcal{S}}\sum_{k=1}^K y_{ik}\log \frac{\exp{(l'_{ik})}}{\sum_{j=1}^{K}\exp{(l'_{ij})}},
\end{equation}

\noindent where $\vl'_i$ denotes the zero-shot normalized logit vector of $\vx_i$, obtained as:

\begin{equation}
\label{eq:zs-normalization}
\vl'_i=\frac{(l^{\text{ZS-max}}_{i}-l^{\text{ZS-min}}_{i})}{(l^{\text{max}}_{i}-l^{\text{min}}_{i})} (\vl_i-l^{\text{min}}_{i}  \mathbf{1})+l^{\text{ZS-min}}_{i} \mathbf{1},
\end{equation}

\noindent with $l^{\text{max}}_{i}=\max_j(l_{ij})$ and $l^{\text{min}}_{i}=\min_j(l_{ij})$, respectively. 
While the calibration strategy formalized in Eq. \ref{eq:ce-logitnorm} forces the \textit{direction} of the logit vector to match the correct category encoded in the one-hot label, its \textit{magnitude} is normalized according to the ZS logit range of image $\vx_i$. Note that this is different from the solution presented in \cite{wei2022mitigating}, as the logit values are normalized by the logit norm, which does not guarantee that the logit values will be in a certain range.

\subsection{Integrating explicit constraints in the learning objective ($\Pen$)}

The problem in \cref{eq:ce-constrained} can also be approximated by an unconstrained problem, for example by transforming the enforced inequality constraints into penalties, which are implemented with the ReLU function. The resulting learning objective can be formally defined as:

\begin{equation}
\label{eq:relu_penalties}
\min_{\boldsymbol{\theta}} \qquad \mathcal{H}(\VY,\VP) + \lambda \sum_{i \in \mathcal{S}} \sum_{k=1}^K (\text{ReLU}(l_{ik}-l^{\text{ZS-max}}_{i})+\text{ReLU}(l^{\text{ZS-min}}_{i}-l_{ik})),
\end{equation}

\noindent where $\lambda$ controls the trade-off between the main loss and the penalties. The intuition behind the penalties is that when the constraint 
in Eq. (\ref{eq:ce-constrained}) is not satisfied, i.e., there exist logit magnitudes outside the zero-shot logit range, the value of the penalty term increases, backpropagating gradients to modify the logit values according to the enforced constraint. We would like to stress that a natural solution to tackle the constrained problem in \cref{eq:relu_penalties} would be the use of Lagrangian multipliers. Nevertheless, as stated earlier, in the context of deep learning, these methods suffer from several well-known limitations, which include training instability and non-convergence due to the difficulty of convexifying loss functions \cite{sangalli2021constrained,birgin2005numerical,dimi96lag}. Thus, despite its simplicity, the use of penalties has proven to be effective in constraining deep models on a myriad of problems, such as image segmentation \cite{kervadec2019constrained}, adversarial attacks \cite{rony2021augmented}, or modeling thermal dynamics \cite{drgovna2021physics}. 

\subsection{Sample-adaptive logit scaling (SaLS)}

Last, we explore a simple but efficient solution that is closely related to temperature scaling (TS) \cite{guo2017calibration}. In particular, TS is a single-parameter variant of Platt scaling \cite{platt1999probabilistic}, which consists in learning the scaling hyperparameter $\tau$ in \cref{eq:clip}. While this strategy has led to very competitive results, it requires an external validation set to fine-tune the value of $\tau$, which limits its use to learning scenarios with abundant labeled data and absence of distributional drifts \cite{ovadia2019can}. Furthermore, $\tau$ is fixed for a whole dataset, which is suboptimal from a sample-wise standpoint. To alleviate these issues, we propose to use the logit normalization defined in \cref{eq:zs-normalization} at inference time to obtain the final softmax probability in \cref{eq:clip}. More concretely, for each sample $i$ to be classified, we compute its zero-shot prediction, whose \textit{min} and \textit{max} logit values are used in \cref{eq:zs-normalization} to scale the logit distribution of that sample $i$ provided by the adaptation method selected. This can be viewed as an \textit{unsupervised sample-wise temperature scaling} during testing, which does not require additional validation samples to fix its value, 
and adapts to the specificity of each sample, regardless of distributional drifts. 


\section{Experiments}
\label{section_experiments}

\subsection{Setup}
\label{subsection_setup}

\paragraph{Datasets.} We use popular datasets for benchmark few-shot \cite{zhou2022coop,yu2023task} and test-time \cite{tpt} CLIP adaptation. \textbf{Domain Generalization:} the adaptation \textit{robustness} to domain shifts is evaluated using ImageNet \cite{deng2009imagenet} distributions. Concretely, we sample a 16-shot training subset from ImageNet's training partition which is directly evaluated on out-of-distribution test data from ImageNetV2 \cite{imagenetV2}, ImageNet-Sketch \cite{imagenetSketch}, ImageNet-A \cite{imagenet_a}, and ImageNet-R \cite{imagenet_r}. \textbf{Fine-grained tasks:} calibration during test-time adaptation is assessed on an assembly of 11 datasets that include heterogeneous discriminative tasks. These include Imagenet \cite{deng2009imagenet}, Caltech101 \cite{caltech}, OxfordPets \cite{oxfordpets}, StanfordCars \cite{stanfordcars}, Flowers102 \cite{flowers102}, Food101 \cite{food101}, FGVCAircraft \cite{aircraft}, SUN397 \cite{sun397}, DTD \cite{dtd}, EuroSAT \cite{eurosat}, and UCF101 \cite{ucf101} datasets. Note that for test time adaptation we uniquely employed their corresponding test partitions. 

\paragraph{Selected methods.} 
Our proposed calibration framework is agnostic to any adaptation strategy of zero-shot models. We evaluate its performance across different popular settings and state-of-the-art methods for CLIP adaptation. \textbf{Prompt Learning (PL):} CoOp \cite{zhou2022coop}, CoCoOp \cite{zhou2022cocoop}, ProGrad \cite{zhu2023prompt} and MaPLe \cite{khattak2023maple} are considered as the baselines. \textbf{Adapters:} CLIP-Adapter \cite{gao2021clip}, TIP-Adapter \cite{zhang2021tip}, and TaskRes\cite{yu2023task} are used. \textbf{Test-Time Adaptation:} TPT~\cite{tpt} is selected as the primary method for test time prompt tuning, together with C-TPT \cite{ctpt}, a concurrent method recently proposed for calibrating TPT. 

\paragraph{CLIP adaptation.} We now describe the experimental details for training the selected adaptation methods. \textbf{Backbones:} All experiments build upon CLIP \cite{radford2021learning}, using its ResNet-50 \cite{resnet} and ViT-B/16 \cite{dosovitskiy2020vit} pre-trained weights. \textbf{Text prompts:} The textual descriptions for zero-shot representation of the target concepts used are the hand-crafted text prompts used in CoOp \cite{zhou2022coop}. \textbf{Image augmentations:} For few-shot adaptation, we applied random zoom, crops, and flips, following \cite{zhou2022cocoop,yu2023task}. Regarding Prompt Learning methods, these transformations are applied continuously during training, while for Adapters, since feature representations are pre-computed, the number of augmentations per support sample is set to 20, following \cite{silva2023closer}. Finally, regarding test-time prompt tuning (TPT), we employed AugMix \cite{augmix} as in \cite{tpt} to form a 64-image batch from each original image. \textbf{Training details:} Adapters are trained following the recent benchmark in \cite{silva2023closer}. We optimized the Adapters for $300$ epochs, using SGD optimizer with a Momentum of $0.9$ and an initial learning rate of $0.1$. In the case of PL, we set the context length of the prompt to 4 and trained CoOp and CoCoOp for 50 and 10 epochs, respectively. We set the same training schedule, optimizer, and learning as in \cite{zhou2022coop}. For ProGrad and MaPLe, we follow the training settings considered for domain generalization reported in theirs respective works \cite{zhu2023prompt,khattak2023maple}. Likewise, for TPT, we optimized the learned prompt by doing a single step with AdamW optimizer, with the learning rate set to 0.005, as in \cite{tpt}.

\paragraph{Evaluation metrics.} To measure the discriminative performance of the different methods, we use classification accuracy (ACC). In terms of calibration, we follow the standard literature and resort to the Expected Calibration Error (ECE). In particular, with $N$ samples grouped into $M$ bins $\{b_1, b_2, \ldots, b_{M}\}$, the ECE is calculated as: $\sum_{m=1}^{M} \frac{\left|b_{m}\right|}{N}\left|\operatorname{acc}\left(b_m\right)-\operatorname{conf}\left(b_m\right)\right|$, where $\operatorname{acc}\left(\cdot\right)$ and $\operatorname{conf}\left(\cdot\right)$ denote the average accuracy and confidence in bin $b_{m}$. 

\paragraph{Calibration details.} We introduced three different alternatives to alleviate the miscalibration of adapted models (\cref{methods}). For $\Norm$ and $\Pen$, we incorporated such modifications during training (\ie adaptation), and kept all implementation details previously presented. Furthermore, the penalty-based calibration weight $\lambda$ in Eq. \ref{eq:relu_penalties} is set to 10 and remains fixed across all settings.

\subsection{Results}
\label{subsection_results}

\begin{table}[b!]
\caption{\textbf{Results for robust Adapters calibration.} The average over the four ImageNet OOD datasets is reported. In brackets, we highlight the difference with respect to each baseline, to stress the impact of the proposed methods ($\Norm$, $\Pen$, and $\Post$). Detailed results are reported in \appen.}
\label{results_calibration_adapters}
\scriptsize
\centering
\subfloat[ResNet-50]{
\begin{tabular}{lcc}
\toprule
\multicolumn{1}{c}{Method}    & \multicolumn{2}{c}{Avg. OOD} \\
\multicolumn{1}{c}{} & ACC   & ECE  \\ \midrule
Zero-Shot \cite{radford2021learning}  & 40.62 & 7.18 \\ \midrule
CLIP-Ad \cite{gao2021clip}            & 34.07 & 15.45 \\
\hspace{1mm} w/ $\Norm$               & 30.06$_{(-4.01)}$\textcolor{red}{$\downarrow$}           & 21.27$_{(+5.82)}$\textcolor{red}{$\uparrow$} \\
\hspace{1mm} w/ $\Pen$                & \textbf{35.20}$_{(+1.13)}$\textcolor{blue}{$\uparrow$}   & 11.22$_{(-4.23)}$\textcolor{blue}{$\downarrow$} \\
\hspace{1mm} w/ $\Post$               & 34.07                                                   & \textbf{8.95}$_{(-6.50)}$\textcolor{blue}{$\downarrow$} \\ \midrule
TIP-Ad(f) \cite{zhang2021tip}         & 41.45 & 19.04 \\
\hspace{1mm} w/ $\Norm$               & 41.73$_{(+0.28)}$\textcolor{blue}{$\uparrow$}            & 19.80$_{(+0.76)}$\textcolor{red}{$\uparrow$} \\
\hspace{1mm} w/ $\Pen$                & \textbf{43.73}$_{(+2.28)}$\textcolor{blue}{$\uparrow$}   & 12.18$_{(-6.86)}$\textcolor{blue}{$\downarrow$} \\
\hspace{1mm} w/ $\Post$               & 41.45                                                   & \textbf{8.13}$_{(-10.91)}$\textcolor{blue}{$\downarrow$} \\ \midrule
TaskRes \cite{yu2023task}             & 41.18 & 11.25 \\
\hspace{1mm} w/ $\Norm$               & \textbf{41.30}$_{(+0.12)}$\textcolor{blue}{$\uparrow$}   & 9.07$_{(-2.18)}$\textcolor{blue}{$\downarrow$} \\
\hspace{1mm} w/ $\Pen$                & 41.29$_{(+0.11)}$\textcolor{blue}{$\uparrow$}            & 10.62$_{(-0.63)}$\textcolor{blue}{$\downarrow$} \\
\hspace{1mm} w/ $\Post$               & 41.18                                                   & \textbf{9.03}$_{(-2.22)}$\textcolor{blue}{$\downarrow$} \\ \midrule
\end{tabular}
}
\subfloat[ViT-B/16]{
\begin{tabular}{lcc}
\toprule
\multicolumn{1}{c}{Method}    & \multicolumn{2}{c}{Avg. OOD} \\
\multicolumn{1}{c}{} & ACC   & ECE  \\ \midrule
Zero-Shot \cite{radford2021learning}  & 57.15 & 4.78 \\ \midrule
CLIP-Ad \cite{gao2021clip}            & 50.61 & 7.82 \\
\hspace{1mm} w/ $\Norm$               & 49.73$_{(+0.88)}$\textcolor{red}{$\downarrow$}         & 12.53$_{(+4.71)}$\textcolor{red}{$\uparrow$} \\
\hspace{1mm} w/ $\Pen$                & \textbf{51.59}$_{(+0.98)}$\textcolor{blue}{$\uparrow$} & 6.38$_{(-1.44)}$\textcolor{blue}{$\downarrow$} \\
\hspace{1mm} w/ $\Post$               & 50.61                                                 & \textbf{4.38}$_{(-3.44)}$\textcolor{blue}{$\downarrow$} \\ \midrule
TIP-Ad(f) \cite{zhang2021tip}         & 25.86 & 63.63 \\
\hspace{1mm} w/ $\Norm$               & 41.64$_{(+15.78)}$\textcolor{blue}{$\uparrow$}          & 58.27$_{(-5.36)}$\textcolor{blue}{$\downarrow$} \\
\hspace{1mm} w/ $\Pen$                & \textbf{49.23}$_{(+23.37)}$\textcolor{blue}{$\uparrow$} & \textbf{40.98}$_{(-22.65)}$\textcolor{blue}{$\downarrow$} \\
\hspace{1mm} w/ $\Post$               & 25.86                                                 & 44.37$_{(-19.26)}$\textcolor{blue}{$\downarrow$} \\ \midrule
TaskRes \cite{yu2023task}             & 58.01 & 7.52 \\
\hspace{1mm} w/ $\Norm$               & \textbf{58.41}$_{(+0.40)}$\textcolor{blue}{$\uparrow$} & \textbf{5.72}$_{(-1.80)}$\textcolor{blue}{$\downarrow$} \\
\hspace{1mm} w/ $\Pen$                & 58.31$_{(+0.30)}$\textcolor{blue}{$\uparrow$}          & 6.65$_{(-0.87)}$\textcolor{blue}{$\downarrow$} \\
\hspace{1mm} w/ $\Post$               & 58.01                                                 & 6.21$_{(-1.31)}$\textcolor{blue}{$\downarrow$} \\ \midrule
\end{tabular}
}
\end{table}

\begin{table}[h!]
\caption{\textbf{Results for robust Prompt Learning calibration.} The average over the four ImageNet OOD datasets is reported. In brackets, we highlight the difference with respect to each baseline, to stress the impact of the proposed methods ($\Norm$, $\Pen$ and $\Post$). Detailed results are reported in \appen.}
\label{results_calibration_pl}
\scriptsize
\centering
\subfloat[ResNet-50]{
\begin{tabular}{lcc}
\toprule
\multicolumn{1}{c}{Method}    & \multicolumn{2}{c}{Avg. OOD} \\
\multicolumn{1}{c}{} & ACC   & ECE  \\ \midrule
Zero-Shot \cite{radford2021learning}  & 40.62 & 7.18 \\ \midrule
CoOp \cite{zhou2022coop}              & 40.86 & 10.97 \\
\hspace{1mm} w/ $\Norm$               & 41.59$_{(+0.73)}$\textcolor{blue}{$\uparrow$} & 10.19$_{(-0.78)}$\textcolor{blue}{$\downarrow$} \\
\hspace{1mm} w/ $\Pen$                & \textbf{41.87}$_{(+1.01)}$\textcolor{blue}{$\uparrow$} & 8.06$_{(-2.91)}$\textcolor{blue}{$\downarrow$} \\
\hspace{1mm} w/ $\Post$               & 40.86                                        & \textbf{7.82}$_{(-3.15)}$\textcolor{blue}{$\downarrow$} \\ \midrule

CoCoOp \cite{zhou2022cocoop}          & 43.36 & 7.69 \\
\hspace{1mm} w/ $\Norm$               & 43.70$_{(+0.34)}$\textcolor{blue}{$\uparrow$} & 7.12$_{(-0.57)}$\textcolor{blue}{$\downarrow$} \\	
\hspace{1mm} w/ $\Pen$                & \textbf{43.86} $_{(+0.50)}$\textcolor{blue}{$\uparrow$} & \textbf{6.15} $_{(-1.54)}$\textcolor{blue}{$\downarrow$} \\
\hspace{1mm} w/ $\Post$               & 43.36                                     & 6.82$_{(-1.87)}$\textcolor{blue}{$\downarrow$} \\ \midrule	
 
ProGrad \cite{zhu2023prompt}          & 42.32 & 7.66 \\
\hspace{1mm} w/ $\Norm$               & 42.21 $_{(+0.11)}$\textcolor{blue}{$\uparrow$} & 7.98 $_{(+0.32)}$\textcolor{red}{$\uparrow$} \\
\hspace{1mm} w/ $\Pen$                & \textbf{42.57}$_{(+0.25)}$\textcolor{blue}{$\uparrow$} & \textbf{6.84}$_{(-0.82)}$\textcolor{blue}{$\downarrow$} \\
\hspace{1mm} w/ $\Post$               & 42.32                                        & 6.90$_{(-0.76)}$\textcolor{blue}{$\downarrow$} \\ \midrule
\end{tabular}
}
\subfloat[ViT-B/16]{
\begin{tabular}{lcc}
\toprule
\multicolumn{1}{c}{Method}    & \multicolumn{2}{c}{Avg. OOD} \\
\multicolumn{1}{c}{} & ACC   & ECE  \\ \midrule
Zero-Shot \cite{radford2021learning}  & 57.15 & 4.78 \\ \midrule
CoOp \cite{zhou2022coop}              & 58.41 & 6.61 \\
\hspace{1mm} w/ $\Norm$               & 58.75$_{(+0.34)}$\textcolor{blue}{$\uparrow$} & \textbf{4.35}$_{(-2.26)}$\textcolor{blue}{$\downarrow$} \\
\hspace{1mm} w/ $\Pen$                & \textbf{59.18}$_{(+0.77)}$\textcolor{blue}{$\uparrow$} & 4.91$_{(-1.70)}$\textcolor{blue}{$\downarrow$} \\
\hspace{1mm} w/ $\Post$               & 58.41                                        & 4.90$_{(-1.71)}$\textcolor{blue}{$\downarrow$} \\ \midrule
CoCoOp \cite{zhou2022cocoop}          & 59.74 & 4.83 \\
\hspace{1mm} w/ $\Norm$               & 59.90$_{(+0.16)}$\textcolor{blue}{$\uparrow$} & 3.94$_{(-0.89)}$\textcolor{blue}{$\downarrow$} \\
\hspace{1mm} w/ $\Pen$                & \textbf{60.20} $_{(+0.46)}$\textcolor{blue}{$\uparrow$} & \textbf{3.89} $_{(-0.94)}$\textcolor{blue}{$\downarrow$} \\
\hspace{1mm} w/ $\Post$               & 59.74                                        & 4.81$_{(-0.00)}$\textcolor{blue}{$\sim$} \\ \midrule
MaPLe \cite{khattak2023maple}          & 60.07 & 4.13 \\
\hspace{1mm} w/ $\Norm$               & 60.09$_{(+0.02)}$\textcolor{blue}{$\uparrow$} & \textbf{3.59}$_{(-0.14)}$\textcolor{blue}{$\downarrow$} \\
\hspace{1mm} w/ $\Pen$                & \textbf{60.62} $_{(+0.55)}$\textcolor{blue}{$\uparrow$} & 3.78 $_{(-0.35)}$\textcolor{blue}{$\downarrow$} \\
\hspace{1mm} w/ $\Post$               & 60.07                                        & 4.38$_{(+0.25)}$\textcolor{red}{$\uparrow$} \\ \midrule
\end{tabular}
}

\end{table}

\paragraph{I) Task 1: Few-shot domain generalization.} Table \ref{results_calibration_adapters} introduces the average few-shot generalization (OOD) results using black-box Adapters, whereas Table \ref{results_calibration_pl} presents the same for PL approaches. We refer the reader to \appen for the detailed results per dataset. First, results consistently show a miscalibration phenomenon when CLIP models are adapted, regardless of the CLIP backbone used, or the transferability approach. \textbf{Few-shot Adapters calibration:} We find that miscalibration is especially occurring in few-shot black-box Adapters. For example, Clip-Ad or TaskRes in \cref{results_calibration_adapters} (a) show ECE increments of $+8.3$ and $+4.0$ respectively. This is further magnified when using the popular TIP-Adapter method. \textbf{Few-shot PL calibration:} PL approaches are relatively more robust in this setting (\eg $+3.8$ CoOp in \cref{results_calibration_pl} (a)). \textbf{On the impact of logit range regularization:} Results show the potential of logit range scaling among its different proposed variants, \textit{improving calibration for all Prompt Learning approaches}, and most of the used Adapters. \textbf{Impact of different strategies to adjust logit range:}. The only strategy that does not allow for consistent performance gains is $\Norm$, which deteriorates performance in some Adapters (see Clip-Ad in Table \ref{results_calibration_adapters}). We believe that the re-parameterization in Eq \ref{eq:ce-logitnorm} might not properly prevent logit range de-adjustment before normalization, and thus overfit to the few support samples. In contrast, $\Pen$ constraint directly regularizes such values, showing consistent ECE decreases for both Adapters (\eg $-22.0$ for TIP-Ad(f) using ViTs, or $-4.3$ for CLIP-Ad using RN50) and PL (\eg $-2.9$ for CoOp using RN50, or $-0.94$ for CoCoOp using ViTs). Interestingly, as a side effect, we also observed accuracy improvements for domain generalization for several methods. Nevertheless, the best calibration performance is provided by a simple, yet effective post-processing standardization, $\Post$. This is especially relevant, since this method \textit{does not require any modification of the adaptation strategy}, and can be potentially applied to the output of any few-shot model.

\paragraph{II) Task 2: Test Time Adaptation (TTA).} 
We report in Table \ref{res:tpt} the performance for test-time prompt tuning across 11 fine-grained adaptation datasets for ResNet-50 backbone. Our results show that compared to zero-shot prediction, TPT largely deteriorates the calibration. Despite this degradation is somehow alleviated by C-TPT, further integrating our approaches show promising potential for better calibration of such methods, with consistent improvements for both strategies (e.g.,$-2.0$ and $-0.9$ in ECE for TPT and C-TPT with $\Post$). 

\setlength{\arrayrulewidth}{0.05mm}

\begin{table}[t!]
\caption{\textbf{Test-time Prompt Learning calibration.} Results for the popular TPT, as well as the concurrent work in \cite{ctpt}, with ResNet-50 backbone, where our three solutions are implemented. Results on ViT-16 backbone are provided in the \appen.}
\label{res:tpt}
\centering
\scriptsize
\begin{tabular}{llrrrrrrrrrrrrr}
\toprule
\multicolumn{1}{l}{} &  & \multicolumn{1}{c}{\textbf{Avg.}} & \multicolumn{1}{c}{\rotatebox{0}{INet}} & \multicolumn{1}{c}{\rotatebox{0}{CAL}} & \multicolumn{1}{c}{\rotatebox{0}{PET}} & \multicolumn{1}{c}{\rotatebox{0}{CAR}} & \multicolumn{1}{c}{\rotatebox{0}{FLW}} & \multicolumn{1}{c}{\rotatebox{0}{FOO}} & \multicolumn{1}{c}{\rotatebox{0}{AIR}} & \multicolumn{1}{c}{\rotatebox{0}{SUN}} & \multicolumn{1}{c}{\rotatebox{0}{DTD}} & \multicolumn{1}{c}{\rotatebox{0}{SAT}} & \multicolumn{1}{c}{\rotatebox{0}{UCF}}   \\
\midrule
\multirow{10}{*}{\rotatebox{90}{ACC}} \hspace{1mm} &  Zero-shot \cite{radford2021learning} & 56.03 & 58.17 & 85.68 & 83.62 & 55.75 & 61.67 & 73.96 & 15.69 & 58.82 & 40.43 & 23.69 & 58.90 \\
\cmidrule{2-14} 
& TPT \cite{tpt} & 58.03 & 60.74 & 87.22 & 84.49 & 58.36 & 62.81 & 74.97 & 17.58 & 61.17 & 42.08 & 28.40 & 60.61 \\
& \hspace{1mm} w/ $\Norm$  & 57.94 & 60.69 & 87.38 & 84.41 & 58.45 & 62.12 & 75.01 & 17.13 & 61.09 & 41.96 & 28.53 & 60.59  \\
& \hspace{1mm} w/ $\Pen$   & 57.69 & 60.74 & 87.06 & 84.30 & 58.13 & 61.84 & 75.17 & 17.22 & 61.11 & 42.02 & 26.60 & 60.35  \\
& \hspace{1mm} w/ $\Post$  & \textbf{58.03} & 60.74 & 87.22 & 84.49 & 58.36 & 62.81 & 74.97 & 17.58 & 61.17 & 42.08 & 28.40 & 60.61  \\
\cmidrule{2-14}
& C-TPT \cite{ctpt}        & 57.54 & 60.02 & 87.18 & 83.65 & 56.41 & 64.80 & 74.89 & 16.62 & 60.72 & 41.55 & 27.06 & 60.01 \\
& \hspace{1mm} w/ $\Norm$  & \textbf{57.63} & 60.00 & 87.06 & 83.65 & 56.57 & 65.04 & 74.82 & 16.86 & 60.58 & 41.61 & 27.51 & 60.27 \\
& \hspace{1mm} w/ $\Pen$   & 57.52 & 60.06 & 86.94 & 83.51 & 56.78 & 64.76 & 74.88 & 16.29 & 60.67 & 41.90 & 26.63 & 60.32 \\ 
& \hspace{1mm} w/ $\Post$  & 57.54 & 60.02 & 87.18 & 83.65 & 56.41 & 64.80 & 74.89 & 16.62 & 60.72 & 41.55 & 27.06 & 60.01 \\
\midrule
\multirow{10}{*}{\rotatebox{90}{ECE}} \hspace{1mm} & Zero-shot \cite{radford2021learning} & 5.04 & 1.90  & 3.56  & 5.64  & 4.17  & 2.10  & 2.35  &  6.31 & 3.79  &  8.60 & 14.40 & 2.66 \\
\cmidrule{2-14}
& TPT \cite{tpt}           & 11.27 & 11.34 & 4.10  & 3.78  & 3.70  & 13.66 & 5.18  & 15.57 & 9.20  & 25.29 & 21.00 & 11.20 \\
& \hspace{1mm} w/ $\Norm$  & 10.57 & 10.81 & 4.29  & 3.71  & 3.62  & 13.29 & 4.73  & 15.28 & 8.50  & 23.95 & 17.61 & 10.49 \\
& \hspace{1mm} w/ $\Pen$   & 9.58 & 11.31 & 3.99  & 1.57  & 2.26  & 13.94 & 4.27  & 14.51 & 8.88  & 23.10 & 11.82 & 9.78 \\
& \hspace{1mm} w/ $\Post$  &  \textbf{9.26} & 9.81  & 4.45  & 2.90  & 2.50  & 12.01 & 3.91  & 15.23 & 8.64  & 21.09 & 12.31 & 9.05 \\
\cmidrule{2-14}
& C-TPT \cite{ctpt}        & 6.33 & 3.05 & 2.60 & 2.46 & 0.87 & 3.91 & 1.62 & 11.30 & 2.73 & 21.38 & 13.58 & 2.88 \\
& \hspace{1mm} w/ $\Norm$  & 5.74 & 2.85 & 2.29 & 2.69 & 0.78 & 3.53 & 1.61 & 10.94 & 2.72 & 20.94 & 12.17 & 2.65\\     
& \hspace{1mm} w/ $\Pen$   & \textbf{3.14} & 5.93 & 2.26 & 2.66 & 0.81 & 3.79 & 1.64 & 11.58 & 2.74 & 20.49 & 10.83 & 2.51 \\     
& \hspace{1mm} w/ $\Post$  & 5.22 & 2.21 & 3.41 & 3.94 & 2.55 & 1.75 & 1.78 & 10.15 & 2.58 & 12.92 & 10.41 & 2.71 \\
\bottomrule
\end{tabular}
\end{table}

\paragraph{III) 
Further constraining the logit range to smaller values.} 
ZS predictions are well calibrated. Nevertheless, during adaptation, the model improves its discriminative performance at the cost of degrading its calibration capabilities. While in this work we advocate for increases of the logit range as a cause of miscalibration, decreasing this range should be done with care. In particular, further decreasing the logit range approaches a scenario of maximum entropy, where the predicted probabilities are semantically meaningless, leading to worse discrimination performance. This reasoning is empirically supported in Table \ref{table:logit-decrease}, where we can see that, regardless of the learning paradigm, significantly decreasing the logit range yields higher ECE scores, i.e., miscalibration is magnified. 

 \begin{table}[h!]
\centering
\caption{\textbf{What if the logit range is further decreased?} ECE scores on ImageNet shifts (V2, S, A and R) for representative methods 
when reducing the original ZS logit range (denoted as $\mathbf{1}$) to half ($\mathbf{1/2}$) and one quarter ($\mathbf{1/4}$) in $\Post$.} 
\setlength{\tabcolsep}{2pt}
\scriptsize
\begin{tabular}{lc|ccc|ccc|ccc}
\toprule
 & \multicolumn{1}{c}{} & \multicolumn{3}{c}{\textcolor{black}{CLIP-Ad}} & \multicolumn{3}{c}{CoOp} & \multicolumn{3}{c}{TPT} \\
 \midrule
\textbf{ZS-Range} & & \textbf{1} & \textbf{1/2} & \textbf{1/4} & \textbf{1} & \textbf{1/2} & \textbf{1/4} & \textbf{1} & \textbf{1/2} & \textbf{1/4} \\
 \midrule
\textbf{RN50} & & 8.95 & 21.31 & 31.51 & 7.82 & 24.44 & 37.72 & 16.74 & 25.15 & 40.7 \\
\bottomrule
\label{table:logit-decrease}
\end{tabular}
\end{table}

\begin{figure}[h!]
    \centering
    \includegraphics[width=0.9\linewidth]{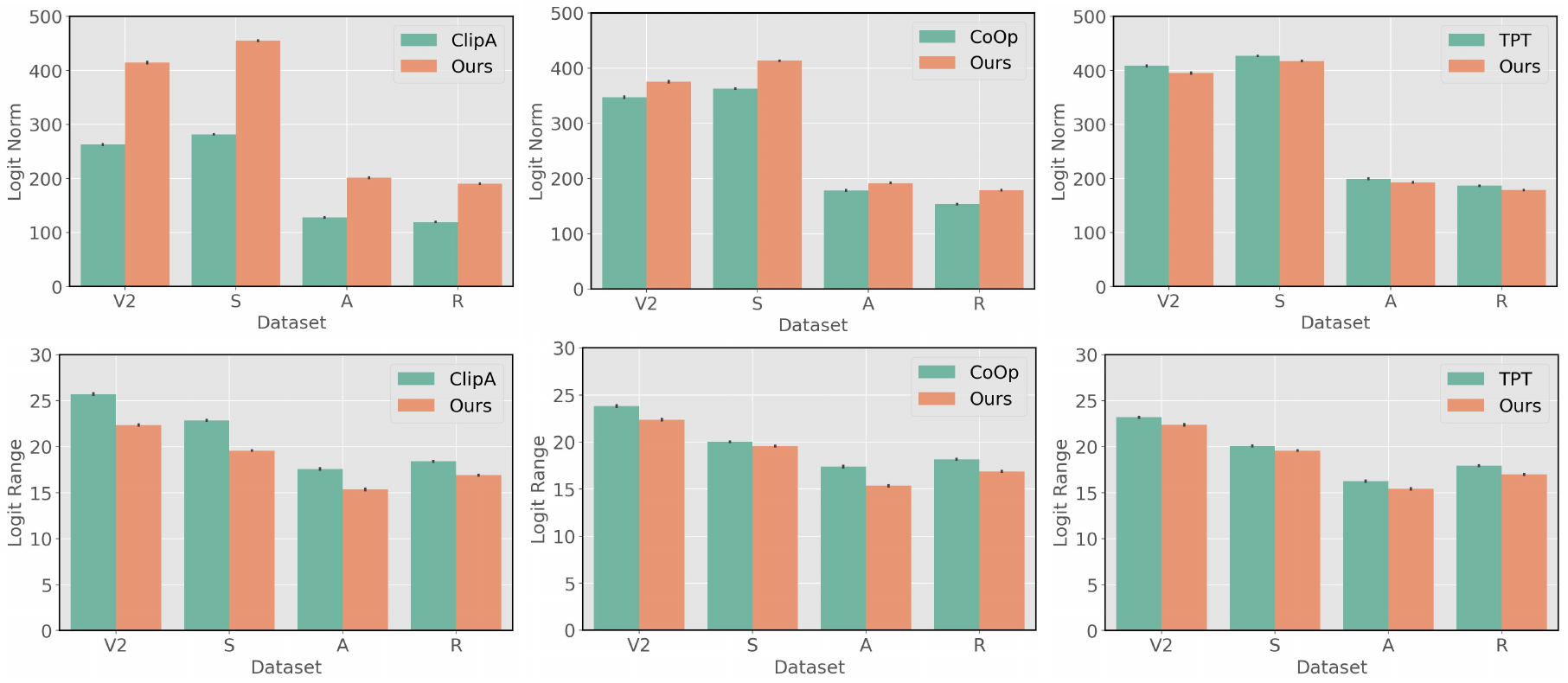}
    \caption{\textbf{Effect of calibrating adapted CLIP models.} Mean of the distribution of logit norms (\textit{top}) and logit ranges (\textit{bottom}) across the four ImageNet OOD datasets for a relevant Adapter-based (CLIP-Ad), Prompt Learning (CoOp) and TPT approach.}
    \label{fig:plotsLogits}
\end{figure}

\paragraph{IV) Effect on logits.} Following one of our main observations (\cref{fig:logit-ranges}), we argued that the source of miscalibration in CLIP adaptation models is the increase of the logit range of their predictions, and not the logit norm. To empirically validate this hypothesis, we depict in \cref{fig:plotsLogits} both the logit norm and logit ranges for a relevant method of each category, as well as the version improved with our $\Post$ solution, across the four OOD datasets of ImageNet. We can observe that, indeed, applying our approach (which improves calibration) leads to reduced logit ranges (\textit{bottom}), whereas the logit norm (\textit{top}) typically increases.

\section{Conclusions}
\label{sec:conclusion}
We have investigated the miscalibration issue of popular CLIP adaptation approaches on the challenging task of few-shot and zero-shot adaptation under distributional drifts. We have analyzed the source of this issue and demonstrated that, in contrast to existing evidence pointing to the logit norm, increases in the range of predicted logits might be a potential cause of miscalibration on the adapted models. To overcome this issue, we have presented three simple solutions, which consist in constraining the logit ranges to the values of the zero-shot predictions, either at training or test time. Extensive experiments on multiple models from the three categories, and popular OOD benchmarks, demonstrate that incorporating our simple solution to existing CLIP adaptation approaches considerably enhances their calibration performance, without sacrificing model accuracy. The proposed approach is model-agnostic, and demonstrate superior performance regardless of the family of approaches or setting, making of our model an appealing yet simple solution for zero-shot and few-shot CLIP adaptation, particularly in the challenging scenario of out-of-distribution data.

\vspace{2mm}

\noindent \textbf{Acknowledgments.} This work is supported by the National Science and Engineering Research Council of Canada (NSERC), via its Discovery Grant program. We also thank Calcul Quebec and Compute Canada.

\bibliographystyle{splncs04}
\bibliography{main}

\clearpage
\setcounter{page}{1}
\resetlinenumber
\section*{Robust Calibration of Large Vision-Language Adapters \\ Supplementary material}
\setcounter{section}{0}

\bigskip
\renewcommand{\thesection}{\Alph{section}}

\section{Supplementary Experiments}

\subsection{Proof of Proposition 1}

If we add a strictly positive constant value $a \in \real_{++}$ to all the elements of a positive logit vector $\vl \geq \mathbf{0}$, the modified vector is $\vl'=\vl + a \mathbf{1}$. Considering $\sigma(\cdot)$ as the softmax function, we can then rewrite the softmax prediction for class $k$ as (we omit the temperature scalar $\tau$ for simplicity, as it does not have any impact on this proof):

\begin{align}
\label{eq:toyexample-softmax}
\sigma_k(\vl')=\sigma_k(\vl+a\mathbf{1}) = \frac{\exp{(l_k + a)}}{\sum_{j=1}^K\exp{(l_j + a)}}&=\frac{\exp{(l_k) \exp{( a)}}}{\sum_{j=1}^K\exp{(l_j)} \exp{(a)}} \nonumber \\
&=\frac{\exp{(a)}}{\exp{(a)}} \frac{\exp{(l_k)}}{\sum_{j=1}^K\exp{(l_j)}} \nonumber \\
&=\frac{\exp{(l_k)}}{\sum_{j=1}^K\exp{(l_j)}}
\end{align}
This proves the first part of Proposition 1.

\noindent \textbf{Showing} $\|\vl'\| \geq \|\vl\|$.

Considering $\|\vl'\|$ as $\|\vl+a\mathbf{1}\|$, we have:

\begin{align}
\label{eq:toyexample-norm}
\|\vl+a\mathbf{1}\|-\|\vl\|&=\sqrt{\sum_{j=1}^K(l_j+a)^2}- \sqrt{\sum_{j=1}^K l_j^2} \nonumber \\
& = \sqrt{\sum_{j=1}^K (l_j^2+2al_j+a^2)} - \sqrt{\sum_{j=1}^K l_j^2} \nonumber \\
& = \sqrt{\sum_{j=1}^K l_j^2+2a\sum_{j=1}^K l_j+ Ka^2)} - \sqrt{\sum_{j=1}^K l_j^2}.
\end{align}

\noindent Since $Ka^2 \geq0$, 
and $2a\sum_{j=1}^K l_j >0$ (we assume $a \in \real_{++}$ and $\vl \geq \mathbf{0}$), 
the first square root is greater than the second one. This results in a positive value for $\|\vl+a\mathbf{1}\|-\|\vl\|$, which demonstrates that $\|\vl+a\mathbf{1}\| = \|\vl'\|\geq \|\vl\|$. 
Thus, Proposition 1 is proved, \textit{i.e., an increase in the logit norm does not necessarily modify the confidence of the predictions.}

\subsection{Proof of Proposition 2}

Considering a scalar $a>1$, $s=a-1$, and $\sigma(\cdot)$ the softmax function, we can define for the predicted class $k (k= \arg \max_j (l_j))$: 

\begin{align}
\label{eq:proof1}
\sigma_k(a\vl)=\frac{e^{al_k}}{\sum_{j=1}^K e^{al_j}}=\frac{e^{(s+1)l_k}}{\sum_{j=1}^K e^{(s+1)l_j}}=\frac{e^{l_k}}{\sum_{j=1}^K e^{l_j+s(l_j-l_k)}}
\end{align}

\noindent where $l_k= \max_j (l_j)$.

If we consider now that for any $j \in [1,2,...,K]$, if $j\neq k$, then $l_j-l_k \leq 0$, we have that:
\begin{align}
l_j + s (l_j - l_k) < l_j \text{ for } j \neq k
\end{align}

Therefore, $e^{l_j + s (l_j - l_k)} < e^{l_j}$ for $j \neq k$ (note that for $j = k$, the exponent remains $l_k$). Thus, the sum in the denominator for $\sigma_k(a \vl)$ is smaller than the sum in the denominator for $\sigma_k(\vl)$:
\begin{align}
\sum_{j=1}^K e^{l_j + s (l_j - l_k)} < \sum_{j=1}^K e^{l_j}
\end{align}

Since the numerator $e^{l_k}$ remains the same, we have:
\begin{align}
\sigma_k(a \vl) = \frac{e^{l_k}}{\sum_{j=1}^K e^{l_j + s (l_j - l_k)}} > \frac{e^{l_k}}{\sum_{j=1}^K e^{l_j}} = \sigma_k(\vl)
\end{align}

This proves that \textit{increasing the logit range (by scaling the logits with a factor $a > 1$) increases the confidence of the predicted class}:
\begin{align}
\sigma_k(a \vl) > \sigma_k(\vl)
\end{align}

\noindent \textbf{Showing} $R(a \vl)>R(\vl)$.

Let us denote the range $R(\vl)$ as:
\begin{align}
R(\vl) = \max(\vl) - \min(\vl)
\end{align}

For a given scalar $a > 1$, we can scale a logit vector $\mathbf{l}$, whose maximum and minimum values are also scaled:

\begin{align}
\max(a \vl) = a \max(\vl) \quad \text{and} \quad \min(a \vl) = a \min(\vl)
\end{align}

Following our definition of range $R(\vl)$:
\begin{align}
R(a \vl) = a \max(\vl) - a \min(\vl)
\end{align}

where $a$ can be factored out, leading to:
\begin{align}
R(a \vl) = a (\max(\vl) - \min(\vl)) = a R(\vl)
\end{align}

Last, as $a>1$, we have that:
\begin{align}
a R(\vl) > R(\vl)
\end{align}

which proves that $R(a \vl) > R(\vl)$. Thus, Proposition 2 is proved.

\subsection{Few shot domain generalization in adapters}
\label{supp:adapters}

We supplement the results depicted in the main manuscript for few-shot adapter calibration (Table \ref{results_calibration_adapters}) by providing results for individual datasets. In particular, we considered in this experiment the popular adapter techniques CLIP-Adpater \cite{gao2021clip}, TIP-Adapter \cite{zhang2021tip}, and TaskRes(e) \cite{yu2023task} and, additionaly, ZS-LP \cite{silva2023closer}. In Table \ref{tab:adapt_full_table}, to each of the above methods, we compare our contributions $\Norm$, $\Pen$, and $\Post$ for ResNet-50 and ViT-B/16 architectures. The adapters are initially trained with source ImageNet dataset, and evaluated under ImageNet distributional shifts, \ie -V2 \cite{imagenetV2}, Sketch \cite{imagenetSketch}, Adversarial \cite{imagenet_a}, and Rendition \cite{imagenet_r}. The classification metric accuracy and calibration metric ECE are reported for the individual datasets. CLIP-Adpater and TIP-Adapter are sensitive to the $\Norm$ technique, which is possible due to the method's dependency on specific hyper-parameter settings \cite{silva2023closer}. Even for these methods, $\Pen$ consistently improves calibration while retaining or improving the accuracy. Last, our post-processing technique $\Post$ can retain the accuracy and assist in calibration consistently across all approaches.

\begin{table}[h!]
\caption{\textbf{Detailed results for robust adapters calibration} Performance of four ImageNet OOD (V2, S, A, R) datasets in different adapters for the proposed $\Norm$, $\Pen$, and $\Post$. These results provide per-dataset performance, and thus complement Table \ref{results_calibration_adapters} in the main manuscript.}
\centering
\scriptsize
\subfloat[][ResNet-50]{
\begin{tabular}{clcccccccccc}
\toprule
 & \multicolumn{1}{c}{Method} & \multicolumn{2}{c}{V2} & \multicolumn{2}{c}{S} & \multicolumn{2}{c}{A} & \multicolumn{2}{c}{R} & \multicolumn{2}{c}{OOD Mean} \\ 
 & & ACC & ECE & ACC & ECE & ACC & ECE & ACC & ECE & ACC & ECE \\ 
\midrule
\multicolumn{1}{c}{\multirow{8}{*}{\rotatebox{90}{}}} & Zero-Shot \cite{radford2021learning} & 51.5 & 3.25 & 33.33 & 3.17 & 21.68 & 21.32 & 55.96 & 0.98 & 40.62 & 7.18\\
 \midrule
  & ZS-LP \cite{silva2023closer} & 51.22 & 11.18 & 27.69	& 16.97 & 17.44	& 34.2	& 48.95 & 11.88 & 36.33	& 18.56\\ 
 & \hspace{1mm} w/ $\Norm$ & 50.95 & 2.78 & 28.87 & 7.12 & 17.48 & 23.72 & 51.37 & 1.81 & \textbf{37.17} & \textbf{8.86} \\  
 & \hspace{1mm} w/ $\Pen$  & 51.38 & 9.31 & 28.36 & 14.42 & 17.43 & 31.79 & 50.69 & 8.21 & \underline{36.97} & 15.93\\
 & \hspace{1mm} w/ $\Post$ & 51.22 & 5.35 & 27.69	& 12.2 & 17.44	& 25.85	& 48.95 & 7.5 & 36.33 & \underline{12.73} \\
 \midrule
 & CLIP-Adapter \cite{gao2021clip} & 49.20 & 9.40 & 25.61 & 13.28 & 15.71 & 31.33 & 45.75 & 7.79 & 34.07 & 15.45\\ 
 & \hspace{1mm} w/  $\Norm$ & 45.83 & 36.47 & 21.65 & 17.28 & 13.69 & 4.09 & 39.08  & 27.23 & 30.06 & 21.27\\ 
 & \hspace{1mm} w/  $\Pen$  & 49.92 & 5.76 & 26.16 & 9.23 & 17.23 & 26.96 & 47.50 & 2.94 & \textbf{35.2} & \underline{11.22} \\ 
 & \hspace{1mm} w/  $\Post$ & 49.20 & 2.64 & 25.61 & 5.27 & 15.71 & 24.42 & 45.75 & 3.45 & \underline{34.07} & \textbf{8.95} \\  
 \midrule
 & TIP-Adapter(f) \cite{zhang2021tip} & 54.37 & 13.46 & 33.58 & 14.49 & 20.79 & 38.13 & 57.07 & 10.08 & 41.45 & 19.04\\
 & \hspace{1mm} w/  $\Norm$ & 53.52 & 15.44 & 34.13 & 15.04 & 21.41 & 38.63 & 57.84 & 10.10 & \underline{41.73} & 19.8\\
 & \hspace{1mm} w/  $\Pen$  & 55.36 & 8.14 & 35.96 & 8.09 & 23.07 & 29.08 & 60.53 & 3.39 & \textbf{43.73} & \underline{12.18}\\ 
 & \hspace{1mm} w/  $\Post$ & 54.37 & 3.09 & 33.58 & 5.43 & 20.79 & 22.51 & 57.07  & 1.49 & 41.45 & \textbf{8.13}\\
 \midrule
 & TaskRes(e) \cite{yu2023task} & 55.07 & 4.91 & 32.66 & 9.13 & 20.33 & 27.52 & 56.67 & 3.42 & 41.18 & 11.25\\
 & \hspace{1mm} w/  $\Norm$& 55.18 & 2.51 & 32.53 & 7.45 & 20.35 & 25.05 & 57.13 & 1.26 & \textbf{41.3}	& \underline{9.07}\\
 & \hspace{1mm} w/  $\Pen$ & 55.23 & 4.06 & 32.83 & 8.46 & 19.63 & 27.77 & 57.48  & 2.17 & \underline{41.29} & 10.62\\ 
 & \hspace{1mm} w/  $\Post$ & 55.07 & 2.46 & 32.66 & 8.24 & 20.33 & 23.43 & 56.67 & 1.98 & 41.18 & \textbf{9.03}\\
  \bottomrule
 \end{tabular}
 }
 \qquad
 \subfloat[][ViT-B/16]{
 \begin{tabular}{clcccccccccccc}
 \toprule
  & \multicolumn{1}{c}{Method} & \multicolumn{2}{c}{V2} & \multicolumn{2}{c}{S} & \multicolumn{2}{c}{A} & \multicolumn{2}{c}{R} & \multicolumn{2}{c}{OOD Mean} \\ 
 & & ACC & ECE & ACC & ECE & ACC & ECE & ACC & ECE & ACC & ECE \\ 
\midrule
\multicolumn{1}{c}{\multirow{8}{*}{\rotatebox{90}{}}} & Zero-Shot \cite{radford2021learning} & 60.71 & 2.37 & 46.18 & 4.76 & 47.73 
 & 8.41 & 73.98 & 3.56 & 57.15 & 4.78\\ \midrule
 & ZS-LP \cite{silva2023closer}   & 60.43 & 10.54 & 41.39 & 17.03 & 42.15 & 20.42 & 70.34 & 3.72 & 53.58 & 12.93\\  
 & \hspace{1mm}  w/ $\Norm$ & 60.20 & 4.26 & 41.81 & 9.52 & 42.48 & 12.90 & 70.85 & 1.63 & \underline{53.84} & \textbf{7.08}\\ 
 & \hspace{1mm}  w/ $\Pen$  &  60.69 & 8.65 & 41.74 & 14.75 & 42.64 & 18.05 & 70.74 & 2.11 & \textbf{53.95} & 10.89\\ 
 & \hspace{1mm}  w/ $\Post$ & 60.43 & 5.99 & 41.39 & 13.19 & 42.15 & 14.92 & 70.34 & 1.15 & 53.58 & \underline{8.81}\\ 
 \midrule
 & CLIP-Adapter \cite{gao2021clip} & 59.88  & 4.77 & 38.73 & 8.86 & 40.13 & 16.20 & 63.71 & 1.43 & 50.61 & 7.82\\  
 & \hspace{1mm}  w/ $\Norm$ & 57.44 & 16.57 & 38.71 & 11.67 & 38.08 & 4.19 & 64.69 & 17.68 & 49.73	& 12.53\\
 & \hspace{1mm}  w/ $\Pen$  & 60.41 & 2.67 & 40.60 & 6.39 & 39.64 & 13.66 & 67.16  & 2.80 & \textbf{51.95} & \underline{6.38} \\ 
 & \hspace{1mm}  w/ $\Post$ & 59.88 & 2.62 & 38.73 & 1.82 & 40.13 & 10.53 & 63.71 & 2.53 & \underline{50.61} & \textbf{4.38} \\
 \midrule
 & TIP-Adapter(f) \cite{zhang2021tip} & 43.50 & 56.17 & 26.64  & 72.79 & 27.29 & 72.24 & 46.01 & 53.31 & 25.86 & 63.63\\
 & \hspace{1mm}  w/ $\Norm$  & 45.68  & 54.26 & 29.33 & 70.57 & 36.17 & 63.74 & 55.38 & 44.49 & \underline{41.64} & 58.27 \\ 
 & \hspace{1mm}  w/ $\Pen$  & 51.78 & 42.77 & 38.44 & 48.24 & 39.15 & 50.67 & 67.55 & 22.22 & \textbf{49.23} & \textbf{40.98}\\
 & \hspace{1mm}  w/ $\Post$ & 43.50 & 42.55 & 26.64 & 50.73 & 27.29 & 50.59 & 46.01 & 33.62 & 35.86 & \underline{44.37}\\
 \midrule
 & TaskRes(e) \cite{yu2023task}  & 64.01 & 4.72 & 45.91 & 10.14 & 46.87 & 14.42 & 75.26 & 0.78 & 58.01 & 7.52\\
 & \hspace{1mm}  w/ $\Norm$ & 64.06 & 2.39 & 46.40 & 7.32 & 47.61 & 11.01 & 75.55 & 2.16 & \textbf{58.41} & \textbf{5.72}\\ 
 & \hspace{1mm} w/ $\Pen$  & 64.17 & 4.03 & 46.36 & 8.85 & 47.39 & 12.57 & 75.31 & 1.16 & \underline{58.31} & 6.65\\ 
 & \hspace{1mm}  w/ $\Post$ & 64.01 & 2.26 & 45.91 & 9.22 & 46.87 & 11.77 & 75.26 & 1.57 & 58.01 & \underline{6.21}\\
 \bottomrule
\end{tabular}
}
\label{tab:adapt_full_table}
\end{table}

\subsection{Few shot domain generalization in prompt learning}

In the following, we extend the evaluation of few-shot prompt learning generalization with per-dataset metrics, and cross-domain generalization.

\paragraph{ImageNet shifts.} In this experiment, prompt learning methods were adapted in 16-shot ImageNet, and evaluated in its corresponding domain drifts (OOD). In this section, we complement the results in Table \ref{results_calibration_pl} with detailed per-dataset metrics and additional prompt learning methods. In particular, we evaluate our proposed calibration methods when applied to CoOp \cite{zhou2022coop}, CoCoOp \cite{zhou2022cocoop}, ProGrad \cite{zhu2023prompt}, and MaPLe \cite{khattak2023maple}. We evaluate CoOp and CoCoOp for both ResNet-50 and ViT-B/16 architectures. As MaPLe is specifically designed for transformer architectures, we consider the ViT-B/16 CLIP backbone. Analogously for ResNet-50, we consider the prompt-aligned gradient technique ProGrad. These results are presented in Table \ref{tab:prompt_full_table}. Following the earlier reported trends, $\Pen$ and $\Post$ consistently provide better calibration and accuracy. In comparison with applying $\Norm$ in Adapters, using them in prompt learning provides stable results, and often provides improved calibration compared to the baseline. It is noteworthy to mention that Prompt learning methods such as CoCoOp, ProGrad, and MaPLe are designed for improved generalization, and thus provide better performance than previously evaluated adapters in Section \ref{supp:adapters}. Despite this, our proposed range re-normalization technique can improve the calibration even for these methods, supporting our observation that the range of the logit indeed plays a key role in calibration. 

\begin{table}[h!]
\caption{\textbf{Detailed results for prompt learning calibration} Performance of four ImageNet OOD (V2, S, A, R) datasets in different prompt learning techniques for the proposed $\Norm$, $\Pen$, and $\Post$. These results provide disentangled results for each task and and serves as a supplement to Table \ref{results_calibration_pl} in the main manuscript.}
\label{tab:prompt_full_table}
\centering
\scriptsize
\subfloat[][ResNet-50]{
\begin{tabular}{clcccccccccccc}
\toprule
 & \multicolumn{1}{c}{Method} & \multicolumn{2}{c}{V2} & \multicolumn{2}{c}{S} & \multicolumn{2}{c}{A} & \multicolumn{2}{c}{R} & \multicolumn{2}{c}{OOD Mean} \\ 
 &  & ACC & ECE & ACC & ECE & ACC & ECE & ACC & ECE & ACC & ECE \\ 
\midrule
\multicolumn{1}{c}{} & Zero-Shot$ $\cite{radford2021learning}  & 51.5 & 3.25 & 33.33 & 3.17 & 21.68 & 21.32 & 55.96 & 0.98 & 40.62 & 7.18 \\
\midrule
 & CoOp \cite{zhou2022coop} & 55.14 & 3.94 & 32.10 & 6.97 & 22.35 & 27.94 & 53.85 & 5.01 & 40.86 & 10.97 \\  
& \hspace{1mm} w/  $\Norm$ & 54.85 & 3.67 & 33.01 & 6.40 & 22.28 & 27.44 & 56.20 & 3.26 & 41.59 & 10.19 \\ 
& \hspace{1mm} w/  $\Pen$  & 55.02 & 2.04 & 34.04 & 3.60 & 22.55 & 24.85 & 55.88 & 1.76 & \textbf{41.87} & \underline{8.06} \\ 
& \hspace{1mm} w/  $\Post$ & 55.14 & 1.54 & 32.10 & 5.68 & 22.35 & 21.95 & 53.85 & 2.11 & \underline{40.86} & \textbf{7.82} \\
 \midrule
 & CoCoOp \cite{zhou2022cocoop} & 55.74 & 2.19 & 35.33 & 3.54 & 23.69 & 23.66 & 58.66 & 1.37 & 43.36 & 7.69 \\  
 &  \hspace{1mm} w/  $\Norm$ & 55.12  & 3.61 & 35.74  & 2.83 & 24.25  & 17.64 & 59.69  & 4.40 & \underline{43.70} & 7.12\\
 &  \hspace{1mm} w/  $\Pen$ & 55.10 & 1.76 & 35.83 & 0.65 & 24.41 & 19.19 & 60.10  & 3.01 & \textbf{43.86} & \textbf{6.15}\\
 &  \hspace{1mm} w/  $\Post$ & 55.74 & 1.40& 35.33 & 4.09 & 23.69 & 21.19 & 58.66 & 0.60 & 43.36 & \underline{6.82} \\
 \midrule
 & ProGrad \cite{zhu2023prompt} & 55.69 & 2.05 & 34.31 & 3.18 & 22.41 & 24.46 & 56.87 & 0.93 & 42.32 & 7.66 \\  
 &  \hspace{1mm}  w/  $\Norm$ & 55.89 & 1.82 & 33.68 & 3.83 & 22.81 & 24.44 & 56.46 & 1.83 & 42.21 & 7.98 \\
 &  \hspace{1mm} w/  $\Pen$  & 55.26 & 1.43 & 34.22 & 2.5 & 23.4 & 22.56 & 57.39 & 0.85 & \textbf{42.57} & \textbf{6.84}\\
 &  \hspace{1mm} w/  $\Post$ & 55.69 & 1.42 & 34.31 & 3.78 & 22.41 & 21.38 & 56.87 & 1.01 & \underline{42.32} & \underline{6.90} \\ 
  \bottomrule
\end{tabular}}
\qquad
\subfloat[][ViT-B/16]{
\begin{tabular}{clcccccccccccc}
\toprule
 & \multicolumn{1}{c}{Method} & \multicolumn{2}{c}{V2} & \multicolumn{2}{c}{S} & \multicolumn{2}{c}{A} & \multicolumn{2}{c}{R} & \multicolumn{2}{c}{OOD Mean} \\ 
 & & ACC & ECE & ACC & ECE & ACC & ECE & ACC & ECE & ACC & ECE \\ 
\midrule
\multicolumn{1}{c}{} & Zero-Shot \cite{radford2021learning} & 60.76 & 2.45 & 46.33 & 7.65 & 47.68 & 8.46 & 74.01 & 3.60 & 57.54 & 6.29 \\
\midrule
 & CoOp \cite{zhou2022coop} & 64.25 & 3.75 & 46.33  & 7.65 & 48.57 & 14.41 & 74.48 & 0.64 & 58.41 & 6.61 \\ 
 &  \hspace{1mm}  w/  $\Norm$ & 64.13 & 1.64 & 47.40 & 2.99 & 48.71 & 10.20 & 74.74  & 2.57 & \underline{58.75} & \textbf{4.35}\\
 &  \hspace{1mm}  w/  $\Pen$  & 64.64 & 1.81 & 47.62 & 4.60 & 49.03 & 11.06 & 75.44 & 2.18 & \textbf{59.18} & 4.91\\
 &  \hspace{1mm} w/  $\Post$ & 64.25 & 1.14 & 46.33 & 5.94 & 48.57 & 9.82 & 74.48 & 2.73 & 58.41 & \underline{4.90} \\
 \midrule
 & CoCoOp \cite{zhou2022cocoop} & 64.24 & 2.18 & 48.45 & 5.00 & 50.12 & 10.99 & 76.16 & 1.14 & 59.74 & 4.83 \\  
 &  \hspace{1mm} w/  $\Norm$ & 64.05 & 2.63 & 48.66 & 0.91 & 50.53 & 6.44 & 76.36 & 5.78 & \underline{59.90} & \underline{3.94}\\
 &  \hspace{1mm}  w/  $\Pen$  & 64.14 & 1.63 & 48.96 & 1.84 & 50.56 & 8.31 & 77.12 & 3.79 & \textbf{60.20} & \textbf{3.89} \\
 & \hspace{1mm} w/  $\Post$ & 64.24 & 1.55 & 48.45 & 5.21 & 50.12 & 9.69 & 76.16 & 2.80 & 59.74 & 4.81\\
 \midrule
 & MaPLe \cite{khattak2023maple} & 64.06	& 1.73 & 48.77 & 3.61 & 50.69 & 8.59 & 76.74 & 2.6 & 60.07 & 4.13 \\  
 &  \hspace{1mm} w/ $\Norm$ &  64.19 & 2.01 & 48.87 & 0.82 & 50.69 & 7.26 & 76.61 & 4.29	 & \underline{60.09} & \textbf{3.59} \\
 &  \hspace{1mm} w/ $\Pen$ &  64.25 & 1.99 & 49.32 & 2.11	& 51.35	& 7.16	& 77.56	& 3.88	& \textbf{60.62}	& \underline{3.78} \\
 &  \hspace{1mm} w/  $\Post$ & 64.06 & 1.57 & 48.77 & 5.05 & 50.69  & 7.69 & 76.74 & 3.19	& 60.07	& 4.38 \\
 \bottomrule
\end{tabular}}
\end{table}

\paragraph{Cross-domain generalization.} This additional experiment evaluates prompt learning methods adapted in ImageNet in 10 fine-grained tasks. These 10 tasks include different target categories, different from the set existing in ImageNet, and evaluate the robustness of the zero-shot capabilities of the learned prompts on new categories. We evaluate the 10 few-shot benchmarks with CoOp based on the prompt learned with the 16-shot setting trained on ImageNet. Results are depicted in Table \ref{tab:coop_cross_domain}. As reported in the literature \cite{zhou2022cocoop}, the prompt learned by CoOp on ImageNet is not sufficient to adapt to the diverse fine-grained tasks, thereby providing lower accuracy and calibration than the original vision-language model (\ie zero-shot). It is worth mentioning that incorporating our logit range normalization techniques, particularly SaLS, provides consistent improvement in calibration compared to the original prompt learning approach CoOp. 

\begin{table}[h!]
\caption{\textbf{Cross-Domain Generalization} Performance of CoOp adapted on ImageNet with 16-shots using ResNet-50, and evaluated on 10 fine-grained tasks. Best results (excluding ZS) in bold.} 
\label{tab:coop_cross_domain}
\vspace{-3mm}
\centering
\scriptsize
\subfloat[][ResNet-50]{
\begin{tabular}{lccccccccccccc}
\toprule
\multicolumn{1}{l}{Method} &                          &  \rotatebox{0}{CAL} & \rotatebox{0}{PET} & \rotatebox{0}{CAR} & \rotatebox{0}{FLW} & \rotatebox{0}{FOO} & \rotatebox{0}{AIR} & \rotatebox{0}{SUN} & \rotatebox{0}{DTD} & \rotatebox{0}{SAT} & \rotatebox{0}{UCF}   & & \textbf{Avg.}  \\
\midrule
Zero-shot \cite{radford2021learning} & \multirow{5}{*}{ACC} &  85.92 & 85.72 & 55.63 & 65.98 & 77.31 & 17.07 & 58.52 & 42.32 & 37.49 & 61.46 & & 58.74 \\
CoOp                       &                      &  85.11 & 84.33 & 52.39 & 57.33 & 72.53 & 13.44 & 55.96 & 33.04 & 24.7 & 56.30 & &  53.51 \\
\hspace{1mm} w/ $\Norm$  &                            &  85.76 & 81.03 & 54.02 & 55.66 & 73.72 & 15.24 & 56.43 & 35.64 & 24.53 & 54.03 & & \underline{53.61} \\
\hspace{1mm} w/ $\Pen$   &                                  & 85.88 & 81.96 & 50.54 & 55.95 & 74.47 & 12.81 & 56.36 & 36.70 & 30.53 & 55.11 & & \textbf{54.03} \\
\hspace{1mm} w/ $\Post$  &                                  & 85.11 & 84.33 & 52.39 & 57.33 & 72.53 & 13.44 & 55.96 & 33.04 & 24.7 & 56.30 & &  53.51 \\
\midrule
Zero-shot \cite{radford2021learning} & \multirow{5}{*}{ECE} &  4.27 & 7.04 & 4.48 & 4.51 & 3.09 & 3.10 & 3.51 & 4.87 & 4.05 & 5.44 & &  4.44\\
CoOp                      &                      & 3.20 & 4.01 & 4.67 & 4.49 & 0.47 & 7.15 & 3.41 & 17.08 & 17.61 & 3.87 & & 6.60 \\
\hspace{1mm} w/ $\Norm$  &  &  2.60 & 3.24 & 7.64 & 5.74 & 1.39 & 5.55 & 2.56 & 15.66 & 10.54 & 5.18 & & 6.01 \\
\hspace{1mm} w/ $\Pen$   &                                  & 3.15 & 3.42 & 7.84 & 5.62 & 4.43 & 6.79 & 2.57 & 12.61 & 8.09 & 3.11 & & \underline{5.76} \\
\hspace{1mm} w/ $\Post$  &                                  &  4.92 & 3.46 & 2.65 & 3.68 & 2.23 & 7.84 & 3.57 & 4.25 & 15.53 & 2.48 & & \textbf{5.06}\\
\bottomrule
\end{tabular}}
\qquad
\subfloat[][ViT-B/16]{
\begin{tabular}{lccccccccccccc}
\toprule
\multicolumn{1}{l}{Method} &                          &  \rotatebox{0}{CAL} & \rotatebox{0}{PET} & \rotatebox{0}{CAR} & \rotatebox{0}{FLW} & \rotatebox{0}{FOO} & \rotatebox{0}{AIR} & \rotatebox{0}{SUN} & \rotatebox{0}{DTD} & \rotatebox{0}{SAT} & \rotatebox{0}{UCF}   & & \textbf{Avg.}  \\
\midrule
Zero-shot \cite{radford2021learning} & \multirow{5}{*}{ACC} &  92.94 & 89.13  & 65.34 & 71.21 & 86.09 & 24.75 & 62.58 & 44.56 & 47.86 & 66.69 & & 65.12 \\
CoOp    &   &   92.29  & 87.60 & 62.85 & 60.66 & 84.50 & 17.40 & 61.14 & 40.37 & 45.93 & 66.38 & & 61.91 \\
\hspace{1mm} w/ $\Norm$  &   &  89.66 & 87.76 & 62.55 & 65.45 & 84.76 & 18.06 & 61.90 & 39.42 & 44.88 & 64.45 & & 61.89\\
\hspace{1mm} w/ $\Pen$   &   & 93.06 & 88.58 & 63.30 & 68.09 & 84.72 & 18.72 & 62.92 & 39.60 & 44.91 & 65.16 & & \textbf{62.91} \\
\hspace{1mm} w/ $\Post$  &   & 92.29 & 87.60 & 62.85 & 60.66 & 84.50 & 17.40 & 61.14 & 40.37 & 45.93 & 66.38 & & \underline{61.91} \\
\midrule
Zero-shot \cite{radford2021learning} & \multirow{5}{*}{ECE} & 5.50 & 4.88 & 4.08 & 4.30 & 2.56 & 3.31 & 1.95 & 3.60 & 4.53 & 2.83 & & 3.75 \\
CoOp &  & 1.93 & 1.54 & 5.43 & 8.05 & 2.56 & 14.39 & 4.65 & 14.07 & 7.40 & 4.78 & & 6.48\\
\hspace{1mm} w/ $\Norm$  &   & 2.83 & 3.68 & 9.59 & 4.52 & 3.42 & 7.89 & 1.88 & 7.60 & 5.27 & 3.65 & & \underline{4.83} \\
\hspace{1mm} w/ $\Pen$   & & 2.76 & 4.10 & 8.48 & 2.16 & 2.16 & 5.41 & 2.19 & 12.97 & 4.41 & 5.65 & & 5.03\\
\hspace{1mm} w/ $\Post$  & & 4.11 & 2.92 & 4.35 & 6.43 & 3.95 & 10.88 & 1.61 & 2.73 & 4.79 & 2.74 & & \textbf{4.45}\\
\bottomrule
\end{tabular}}
\end{table}

\subsection{Test time prompt tuning with ImageNet OOD benchmark}
Test time prompt tuning methods, such as TPT \cite{tpt}, provide a provision to infer an individual sample directly during the test time. In this supplementary experiment, we analyze our methods with TPT for the ImageNet OOD datasets and complement results on fine-grained datasets depicted in Table \ref{res:tpt}. The numbers for each ImageNet OOD dataset comparing TPT with our methods are reported in Table \ref{tab:tpt_ood_full_table}. In this setting, Zero-Shot inference is better calibrated than the TPT, even when the accuracy increases with adaptation. This drastic degradation in calibration may be largely due to the use of entropy, which favours larger distances between the winner and other logits, thereby increasing the logit range. Through our methods, we have attempted to restrict the logit range from going beyond the Zero-Shot range, providing us the expected improvement in calibration, and even in accuracy in some cases. 

\begin{table}[h!]
\caption{\textbf{Additional tasks for test time prompt tuning calibration} Performance of ImageNet OOD datasets (V2, S, A, R) with TPT for the proposed $\Norm$, $\Pen$, and $\Post$. These results supplement the ones depicted in Table \ref{res:tpt} in the main manuscript by integrating four more adaptation tasks.}
\label{tab:tpt_ood_full_table}
\centering
\scriptsize
\subfloat[][ResNet-50]{
\begin{tabular}{clcccccccccccc}
\toprule
 & \multicolumn{1}{c}{Method} & \multicolumn{2}{c}{V2} & \multicolumn{2}{c}{S} & \multicolumn{2}{c}{A} & \multicolumn{2}{c}{R} & \multicolumn{2}{c}{OOD Mean} \\ 
 & & ACC & ECE & ACC & ECE & ACC & ECE & ACC & ECE & ACC & ECE \\ 
\midrule
\multicolumn{1}{c}{\multirow{6}{*}{\rotatebox{90}{}}} & Zero-Shot \cite{radford2021learning}  & 51.50 & 3.25 & 33.33 & 3.17 & 21.68 & 21.32 & 55.96 & 0.98 & 40.62 & 7.18 \\
 & TPT \cite{tpt} & 54.97 & 13.77 & 35.03 & 15.28 & 26.61 & 30.82 & 59.00 & 10.45 & 43.90 & 17.58 \\ 
 & \hspace{1mm} w/ $\Norm$  & 54.91 & 13.18 & 35.02 & 14.53 & 26.65 & 29.49 & 59.01 & 9.78 & \textbf{43.90} & \underline{16.75} \\
  & \hspace{1mm} w/ $\Pen$ & 54.87 & 13.65 & 35.02 & 15.22 &  26.17 & 30.50 & 58.86 & 9.99 & 43.73 & 17.32\\
  & \hspace{1mm} w/ $\Post$ & 54.97 & 12.15 & 35.03 & 13.72 & 26.61 & 27.98 & 59.00 & 7.77 & \underline{43.89} & \textbf{16.74} \\
 \bottomrule
 \end{tabular}
 }
 \qquad
 \subfloat[][ViT-B/16]{
 \begin{tabular}{clcccccccccccc}
 \toprule
 & \multicolumn{1}{c}{Method} & \multicolumn{2}{c}{V2} & \multicolumn{2}{c}{S} & \multicolumn{2}{c}{A} & \multicolumn{2}{c}{R} & \multicolumn{2}{c}{OOD Mean} \\ 
 & & ACC & ECE & ACC & ECE & ACC & ECE & ACC & ECE & ACC & ECE \\ 
 \midrule
\multicolumn{1}{c}{\multirow{6}{*}{\rotatebox{90}{}}} & Zero-Shot \cite{radford2021learning} & 60.83 & 2.40 & 46.15 & 4.80  & 47.80 & 8.36 & 73.99 & 3.51 & 57.19 & 4.77\\
 & TPT \cite{tpt} & 63.69 & 11.61 & 47.91 & 16.16 & 54.84 & 14.68 & 77.13 & 4.77 & 60.89	& 11.81\\  
 & \hspace{1mm} w/ $\Norm$ & 63.56 & 11.14 & 47.76 & 15.71 & 54.57 & 13.60 & 77.08 & 4.60 & 60.74 & \underline{11.26}\\ 
  & \hspace{1mm} w/ $\Pen$ & 63.53 & 11.59 & 47.94  & 16.00 & 54.55 & 13.87 &  77.07 & 4.14 & \underline{60.77} & 11.40\\
 & \hspace{1mm} w/ $\Post$ & 63.69 & 10.48 & 47.91 & 15.70 & 54.84 & 13.52 & 77.13 & 3.75 & \textbf{60.89} & \textbf{10.86}\\
 \bottomrule
\end{tabular}}
\end{table}

\subsection{Additional experiments for Test time prompt tuning}
In this section, we further study TPT with our methods on 11 few-shot benchmarks. In particular, we complement the results provided in the main manuscript (Table \ref{res:tpt}) with ImageNet results and the CLIP ViT-B/16 model. In \cref{tab:tpt_fs_full_table}, the overall (Avg.) results show that our calibration methods are better than the baselines, especially in calibration. As expected from a good post-processing technique, $\Post$ retains the accuracy and consistently improves the calibration across tasks. Importantly, even for C-TPT, our method $\Post$ still improves the calibration, proving that even with the best prompt choice for calibration, there is still scope for improvement by adjusting the predictions logit range. More importantly, our approach $\Post$ can be directly applied to the logit predictions, not requiring pre-training the network, such as C-TPT, making of it an efficient \textit{ready-to-use} solution.

\begin{table}[h!]
\caption{\textbf{Additional results for Test time prompt tuning calibration for ViT-B/16 backbone} Performance of popular 11 few shot datasets with TPT for the proposed $\Norm$, $\Pen$, and $\Post$. Best result over TPT are highlighted in bold, and second underlined.}
\label{tab:tpt_fs_full_table}
\centering
\scriptsize
\begin{tabular}{llrrrrrrrrrrrrr}
\toprule
\multicolumn{1}{l}{} &  & \multicolumn{1}{c}{\textbf{Avg.}} & \multicolumn{1}{c}{\rotatebox{0}{INet}} & \multicolumn{1}{c}{\rotatebox{0}{CAL}} & \multicolumn{1}{c}{\rotatebox{0}{PET}} & \multicolumn{1}{c}{\rotatebox{0}{CAR}} & \multicolumn{1}{c}{\rotatebox{0}{FLW}} & \multicolumn{1}{c}{\rotatebox{0}{FOO}} & \multicolumn{1}{c}{\rotatebox{0}{AIR}} & \multicolumn{1}{c}{\rotatebox{0}{SUN}} & \multicolumn{1}{c}{\rotatebox{0}{DTD}} & \multicolumn{1}{c}{\rotatebox{0}{SAT}} & \multicolumn{1}{c}{\rotatebox{0}{UCF}}   \\
\midrule
\multirow{10}{*}{\rotatebox{90}{ACC}} \hspace{1mm} &  Zero-shot \cite{radford2021learning} & 63.92 & 66.72 & 93.31 & 88.25 & 65.51 & 67.40 & 83.64 & 23.91 & 62.56 & 44.39 & 42.22 & 65.24 \\
\cmidrule{2-14} 
& TPT \cite{tpt} &  65.18  & 68.87 & 94.28 & 87.41 & 66.51 & 68.98 & 84.64 & 23.43 & 65.61 & 46.69 & 42.49 & 68.12 \\
& \hspace{1mm} w/ $\Norm$  &  65.13  & 68.83 & 94.16 & 87.54 & 66.61 & 68.66 & 84.67 & 23.31 & 65.39 & 46.34 & 42.98 & 67.99 \\
& \hspace{1mm} w/ $\Pen$ & \textbf{65.22}  & 68.86 &  94.08 & 86.64 & 66.72 & 68.82 & 84.43 & 23.07 & 65.55 & 46.28 & 44.99 & 67.94 \\
& \hspace{1mm} w/ $\Post$  & \underline{65.18}  & 68.87 & 94.28 & 87.41 & 66.51 & 68.98 & 84.64 & 23.43 & 65.61 & 46.69 & 42.49 & 68.12 \\
\cmidrule{2-14}
& C-TPT \cite{ctpt}  & 64.59 & 68.08 & 93.63 & 88.20 & 65.75 & 69.43 & 83.07 & 24.03 & 64.52 & 46.16 & 42.20 & 65.42 \\
& \hspace{1mm} w/ $\Norm$  & 64.41 & 68.09 & 93.79 & 88.28 & 65.87 & 69.27 & 83.05 & 23.91 & 64.30 & 45.63 & 41.28 & 65.05  \\
& \hspace{1mm} w/ $\Pen$  & \textbf{64.68} & 68.00 & 93.39 & 88.06 & 65.81 & 69.27 & 83.16 & 24.39 & 64.59 & 45.69 & 43.96 & 65.11 \\
& \hspace{1mm} w/ $\Post$  & \underline{64.59} & 68.08 & 93.63 & 88.20 & 65.75 & 69.43 & 83.07 & 24.03 & 64.52 & 46.16 & 42.20 & 65.42 \\
\midrule
\multirow{10}{*}{\rotatebox{90}{ECE}} \hspace{1mm} & Zero-shot \cite{radford2021learning} & 3.91 & 1.86 & 5.08	& 4.19 & 4.22 & 1.87 & 1.79 & 5.21 & 1.96 & 7.87 & 6.52 & 2.50 \\
\cmidrule{2-14}
& TPT \cite{tpt}  & 11.36  & 10.42 & 4.41 & 5.45 & 5.11 & 13.13 & 4.24 & 16.76 & 11.26 & 21.23 & 20.42 & 11.54 \\
& \hspace{1mm} w/ $\Norm$ & 10.82 & 10.21 & 4.22 & 5.27 & 4.86 & 12.93 & 3.98 & 16.35 & 10.58 & 20.92 & 18.28 & 11.38\\
& \hspace{1mm} w/ $\Pen$  & \textbf{9.27}  & 10.23 & 4.26 & 3.44 & 3.63 & 12.33 & 3.25 & 15.77 & 10.85 & 19.62 & 8.53 & 10.07 \\
& \hspace{1mm} w/ $\Post$  & \underline{9.88} & 9.56 & 4.55 & 5.00 & 3.86 & 11.50 & 4.30 & 15.54 & 10.89 & 18.90 & 13.88 & 10.69\\
\cmidrule{2-14}
& C-TPT \cite{ctpt} & 4.81 & 3.00 & 4.12 & 1.46 & 1.35 & 5.29 & 2.66 & 4.11 & 5.06 & 12.41 & 11.33 & 2.16 \\
& \hspace{1mm} w/ $\Norm$  & 5.03  & 3.01 & 4.36 & 1.43 & 1.51 & 5.35 & 2.71 & 4.25 & 4.74 & 12.24 & 11.13 & 2.53 \\
& \hspace{1mm} w/ $\Pen$   & \underline{4.61} & 3.09 & 3.76 & 1.60 & 1.20 & 5.46 & 2.59 & 4.25 & 4.87 & 13.71 & 6.81 & 1.86 \\
& \hspace{1mm} w/ $\Post$ & \textbf{4.48} & 2.22 & 4.38 & 3.61 & 2.53 & 2.29 & 1.45 & 5.60 & 3.32 & 9.2 & 10.46 & 4.20 \\
\bottomrule
\end{tabular}
\end{table}

\subsection{Additional studies on Logit norm, range, and calibration}
\begin{figure}[h!]
    \centering
    \includegraphics[width=\linewidth]{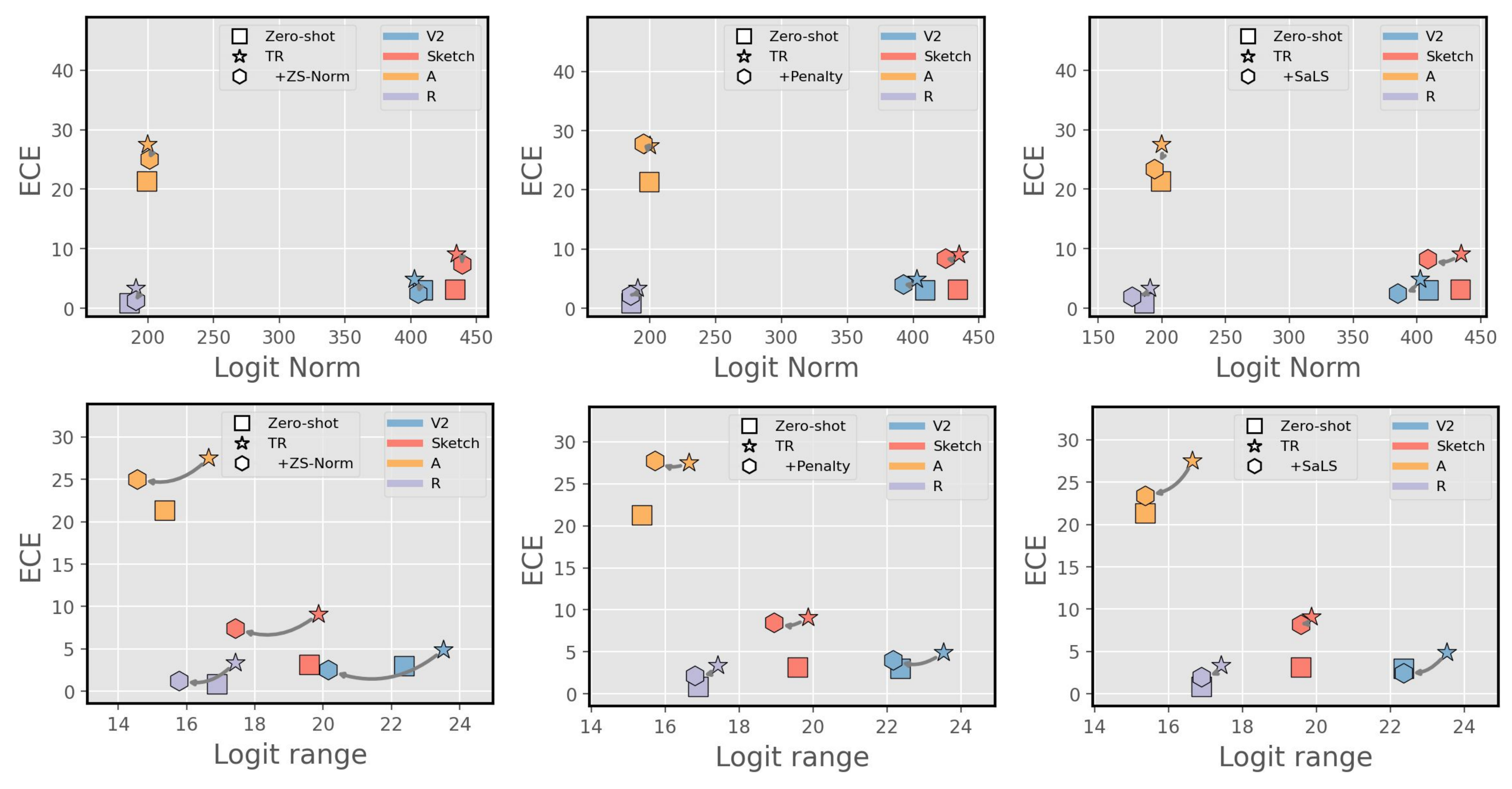}
    \caption{\textbf{Additional Logit studies for few-shot Adapters.} Analysis of average Logit norm and range after improving the calibration of the Adapter model TaskRes \cite{yu2023task} using the proposed logit range regularization methods for improved calibration, i.e., $\Norm$ (\textit{left}), $\Pen$ (\textit{middle}) and $\Post$ (\textit{right}).}
    \label{fig:tr-logit-ranges}
\end{figure}

\begin{figure}[h!]
    \centering
    \includegraphics[width=\linewidth]{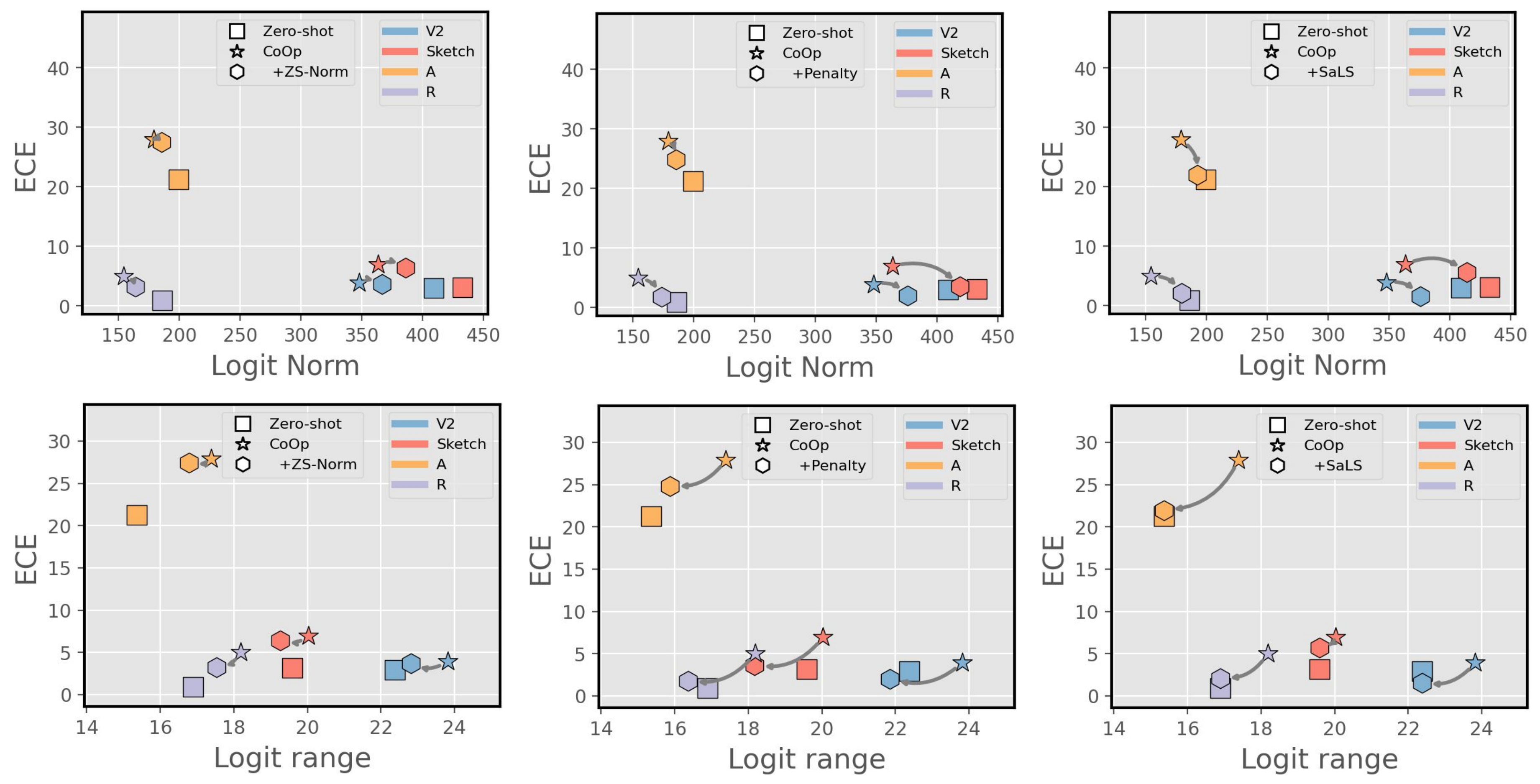}
    \caption{\textbf{Additional Logit studies for Prompt Learning.} Analysis of average Logit norm and range after improving the calibration of the Prompt learning technique CoOp \cite{zhou2022coop} using the proposed logit range regularization methods for improved calibration, i.e., $\Norm$ (\textit{left}), $\Pen$ (\textit{middle}) and $\Post$ (\textit{right}).}
    \label{fig:coop-logit-ranges}
\end{figure}

In this experiment, we analyze the impact of our contributions in calibration to the logit norm and range. We consider representative methods for few-shot Adapters and Prompt Learning, \ie, TaskRes \cite{yu2023task} and CoOp \cite{zhou2022coop}, respectively. Fig. \ref{fig:tr-logit-ranges}, and \ref{fig:coop-logit-ranges} depict the comparison of logit norm/range with ECE for $\Norm$, $\Pen$, and $\Post$ proposed calibration methods. As expected, after applying our method, ECE is reduced in most scenarios. It is worth mentioning that ECE improvements correlate with the decrease in the logit range. This is not the case of the logit norm, which either increases or remains constant. These observations correlate with our hypothesis in the main paper, and demonstrate that logit range plays a key role in calibration. 

\begin{table}[h!]
\caption{\textbf{Comparison to unsupervised calibration approaches.} Average results are reported on ImageNet OOD.}
\centering
\setlength{\tabcolsep}{1.5pt}
\scriptsize
\label{table:add-calibration}
\begin{tabular}{lcc|cc|cc}
\midrule
 & \multicolumn{2}{c}{\textbf{+SaLS}} & \multicolumn{2}{c}{\textbf{+L-Norm \cite{wei2022mitigating}}} & \multicolumn{2}{c}{\textbf{+ECP} \cite{pereyra2017regularizing}} \\
 \midrule
  & \textbf{ACC} & \textbf{ECE} & \textbf{ACC} & \textbf{ECE} & \textbf{ACC} & \textbf{ECE} \\
 \midrule
\textbf{TaskRes}$_{\text{\tiny{CVPR'23}}}$ &  41.18 & 9.03 & 42.04 & 14.40 & 41.37 & 9.50 \\ 
 \textbf{CoOp}$_{\text{\tiny{IJCV'22}}}$ &  40.86 & 7.82 & 40.82 & 18.66 & 41.81 & 23.76 \\ 
 \textbf{TPT}$_{\text{\tiny{NeurIPS'22}}}$ & 58.77 & 9.21 & 57.46 & 10.91 & 57.80 & 11.32 \\ 
\midrule 
\end{tabular}
\end{table}

\subsection{Comparison to other calibration methods}

We further evaluate the performance of our simplest solution, $\Post$, compared to several existing unsupervised calibration approaches. Our reasoning behind using these methods, i.e., L-Norm \cite{wei2022mitigating} and ECP \cite{pereyra2017regularizing}, stems from the fact that they do not require labeled samples, in contrast to most existing methods (for example, Temperature Scaling (TS) needs a large validation set to fix the temperature value). These results, which are reported in Table \ref{table:add-calibration}, showcase that the proposed post-processing alternative brings important performance gains, in terms of calibration, without sacrificing discriminative power. This gap is particularly significant in prompt learning, where the proposed $\Post$ improves the ECE on CoOp by 11\% and 16\%, compared to L-Norm and ECP, respectively.

\subsection{Reliability plots}
\begin{figure}[h!]
    \centering
    \includegraphics[width=\linewidth]{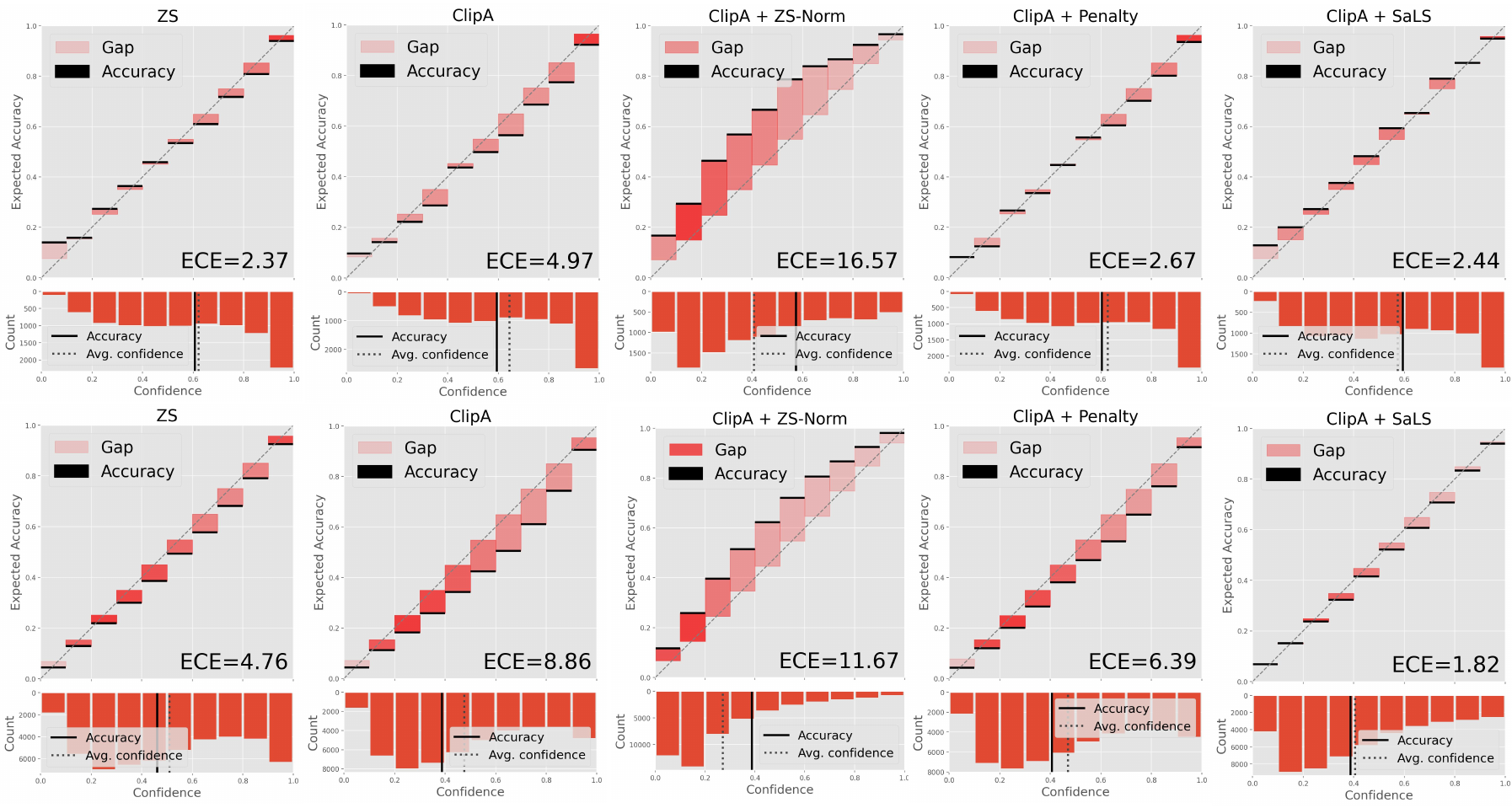}
    \caption{Reliability plot for Adapter method Clip-Adapter with ImageNet Variants, ImageNetV2 (Top), and ImageNetSketch (Bottom)}
    \label{fig:rp_clipa}
\end{figure}

\begin{figure}[h!]
    \centering
    \includegraphics[width=\linewidth]{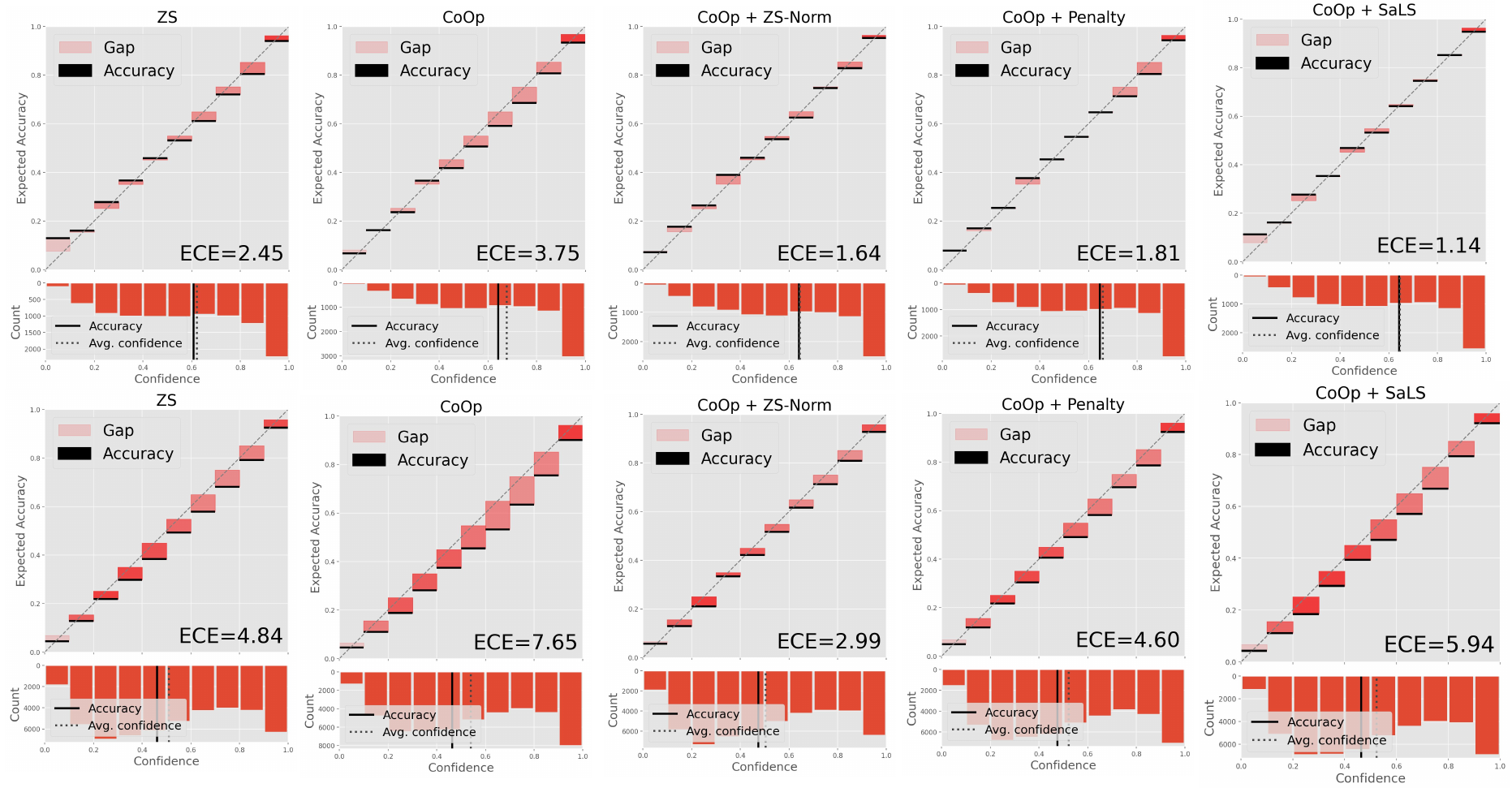}
    \caption{Reliability plot for Prompt learning method CoOp with ImageNet Variants, ImageNetV2 (Top), and ImageNetSketch (Bottom)}
    \label{fig:rp_coop}
\end{figure}

\begin{figure}[h!]
    \centering
    \includegraphics[width=\linewidth]{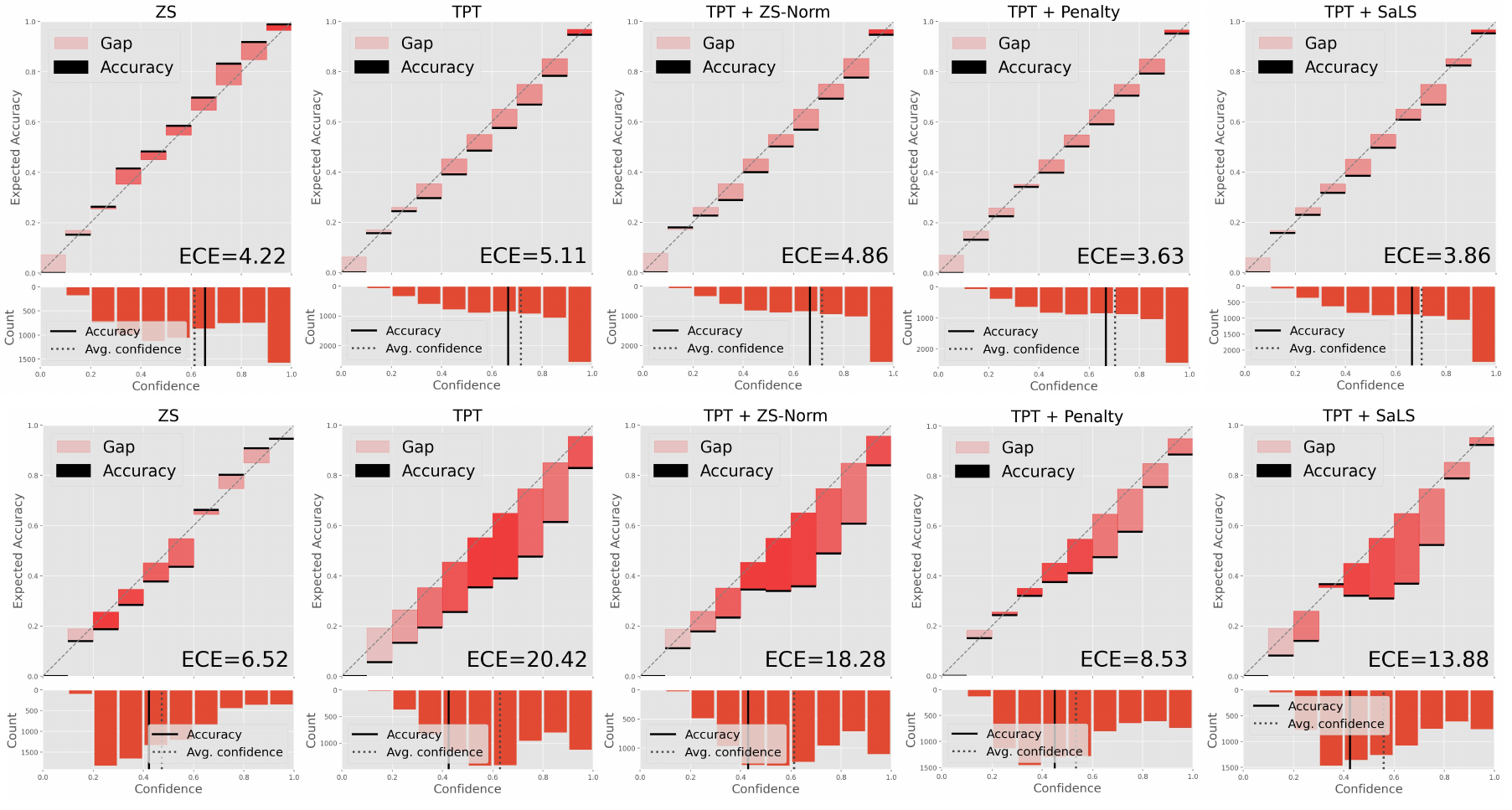}
    \caption{TPT few shot benchmark reliability plot comparison for ViT-B/16 architecture. From Top to bottom: StanfordCars, EuroSAT}
    \label{fig:rp_tpt1}
\end{figure}

Fig. \ref{fig:rp_clipa}, \ref{fig:rp_coop}, and \ref{fig:rp_tpt1} depict the reliability plot of ZS, $\Norm$, $\Pen$ and $\Post$ for one from each of the setting of Adapters (Clip-Adapter), Prompt learning (CoOp), and Test time prompt tuning (TPT) for few representative cases in ImageNet OOD, and Few shot benchmarks respectively. From these plots, it could be noted that the density of the plots near the expected calibration curve is lower in our methods compared to the baselines, moving closer to ZS without compromising much the accuracy. 
Furthermore, the difference between the accuracy and the average confidence (in the bottom of the plots) is typically reduced when our approaches are integrated into the original methods, a sign that indicates that a model is better calibrated.

\end{document}